\documentclass[journal]{IEEEtran}

\usepackage{graphicx}
\usepackage{rotating}
\usepackage{booktabs}
\usepackage{amsmath,amssymb}
\usepackage{cite}
\usepackage{url}
\usepackage{array}
\usepackage{microtype}

\usepackage{tabularx}

\usepackage[table]{xcolor}

\definecolor{revisionblue}{RGB}{0,0,0}

\newcommand{\reviewgreen}[1]{#1}

\newcommand{\reviewblue}[1]{#1}

\usepackage{xurl}
\usepackage{hyperref}

\DeclareUnicodeCharacter{2265}{\ensuremath{\geq}}
\DeclareUnicodeCharacter{2264}{\ensuremath{\leq}}
\DeclareUnicodeCharacter{00B1}{\ensuremath{\pm}}
\DeclareUnicodeCharacter{00D7}{\ensuremath{\times}}
\DeclareUnicodeCharacter{2212}{-}

\usepackage{tikz}
\usetikzlibrary{arrows.meta,positioning}

\graphicspath{{./}{figures/}}
\newcommand{\NRaw}{16,916}
\newcommand{\NDedup}{9,843}
\newcommand{\NHighConf}{400}
\newcommand{\NBoundary}{50}
\newcommand{\NFullTextExcluded}{\reviewblue{122}}

\newcommand{\NAnalytical}{\reviewblue{278}}

\newcommand{\NRisk}{\reviewblue{7}}

\newcommand{\NCompleteChains}{\reviewblue{31}}
\newcommand{\NCompleteChainsFive}{\reviewgreen{110}}

\newcommand{\NMTEGNodes}{\reviewblue{359}}
\newcommand{\NMTEGEdges}{\reviewblue{3,284}}
\newcommand{\PExternal}{\reviewblue{28.4\%}}
\newcommand{\PCalibration}{\reviewblue{20.1\%}}
\newcommand{\PDCA}{\reviewblue{13.3\%}}
\newcommand{\PGenomics}{\reviewblue{6.5\%}}
\newcommand{\PTranscriptomics}{\reviewblue{2.2\%}}
\newcommand{\PMultiomics}{\reviewblue{4.0\%}}
\newcommand{\PProteomics}{\reviewblue{0.4\%}}
\newcommand{\PLongitudinal}{\reviewblue{17.3\%}}

\newcommand{\TaxonomyCoverage}{\reviewblue{85.7\%}}
\newcommand{\IntegrationDensity}{\reviewblue{0.565217}}
\newcommand{\CrossCommunityMixing}{\reviewblue{0.277}}
\newcommand{\FragmentationScore}{\reviewblue{0.386}}

\newif\ifHumanReviewCompleted
\HumanReviewCompletedtrue

\title{Disentangling Lung-Cancer CT/LDCT AI: A Systematic Evidence Map of Clinical Tasks, Evidence Chains, and Translational Gaps}

\author{Surajit Das
\thanks{Surajit Das is the corresponding author. Surajit Das: mr.surajitdas@gmail.com}}

\newif\ifAnonymousReview
\AnonymousReviewtrue
\newcommand{\ReviewRepository}{\url{ https://drive.go
ogle.com/drive/folders/1BX1b6pcbVJ-vEkQl2sSuwTvlvaHZ
-YU?usp=sharing}}
\newcommand{\ReviewRegistration}{\ifAnonymousReview registration identifier withheld for anonymous review\else\url{https://osf.io/2nyme/}\fi}
\begin{document}
\maketitle

\begin{abstract}
\textbf{Background:} Artificial-intelligence studies using computed tomography (CT) for lung cancer are often broadly labelled ``prediction'' despite addressing clinically distinct tasks. \textbf{Methods:} We systematically mapped CT/low-dose CT (LDCT)-centered lung-cancer AI using five-database retrieval, full-text eligibility assessment, role-aware modality/omics extraction, clinical-task classification, and a Multi-Tier Evidence Graph (MTEG). \textbf{Results:} The final corpus comprised \reviewblue{278 studies (2016--2026): 216 Detection, 7 future Risk-prediction, and 55 Other studies}. Clinical variables (\reviewblue{96.0\%}), 3-D CT/LDCT (\reviewblue{74.5\%}), and radiomics (\reviewblue{64.0\%}) predominated, whereas external validation (\reviewblue{28.4\%}), calibration (\reviewblue{20.1\%}), decision-curve analysis (\reviewblue{13.3\%}), longitudinal CT (\reviewblue{17.3\%}), and saliency/attribution XAI (\reviewblue{21.6\%}) were less frequent. The fixed-extraction MTEG comprised 359 nodes and 3,284 edges. Of 278 studies, 31 (11.15\%) covered all six substantive tiers, with the greatest attrition at reasoning/explanation. With explanation optional, 110 studies (39.6\%) covered the remaining five tiers. \textbf{Conclusions:} Overall, the literature is detection-dominated, genuine future risk prediction remains uncommon, and complete translational evidence chains are rare in the extracted map. Limited omics-role recall constrains interpretation of the mapped molecular-input frequencies.
\end{abstract}

\begin{IEEEkeywords}
lung cancer, computed tomography, low-dose CT, artificial intelligence, radiomics, risk prediction, pulmonary nodule, systematic evidence mapping, complex networks, semantic networks, evidence graph, genomics, multi-omics, clinical validation.
\end{IEEEkeywords}

\section{Introduction}
\IEEEPARstart{L}{ung} cancer screening, pulmonary-nodule assessment, and thoracic imaging have become major targets for machine learning because CT and low-dose CT (LDCT) provide rich image phenotypes that can be combined with clinical risk factors, radiomic descriptors, longitudinal change, and molecular information. However, growth in the number of published models does not necessarily indicate progress in translational maturity. A literature can be algorithmically rich while remaining clinically fragmented if model development is disconnected from external validation, calibration, decision analysis, explicit reasoning, uncertainty, or clinically actionable output.

{These observations motivated three interrelated challenges that emerged from the review scope and preliminary mapping of the CT/LDCT lung-cancer AI literature: ambiguity in the clinical tasks grouped under ``prediction,'' inconsistency in what is described as multimodal evidence, and limited visibility of whether reported methodological components connect into a coherent translational pathway. Together, these challenges affect how reliably the literature can be synthesized, compared, and interpreted in terms of clinical translation.}

A first challenge is \emph{clinical-task heterogeneity}. \reviewblue{\cite{REC_06514,REC_06367,REC_06385,REC_06613,REC_08222,REC_05321,REC_05363,REC_05501,REC_05652,REC_06035}} Estimating future cancer incidence in an asymptomatic screening population differs fundamentally from classifying whether an already detected pulmonary nodule is malignant, and both differ from predicting treatment response, survival, molecular mutation status, pulmonary function, or other disease-related outcomes. These tasks may use overlapping language such as ``prediction,'' ``risk,'' or ``classification,'' but they operate at different clinical decision points and imply different validation requirements. Accordingly, a review that pools these tasks without explicit classification may produce misleading estimates of modality use, model performance, validation practice, and clinical readiness. This distinction is increasingly important as reporting and risk-of-bias guidance for prediction models places greater emphasis on transparent development, evaluation, calibration, and applicability \cite{tripodai,probastai}, while medical-AI guidance emphasizes clinical evaluation and human-centered deployment \cite{claim,decideai}.

A second challenge concerns the meaning of \emph{multimodal} (interpretation of multimodal evidence). \reviewblue{\cite{REC_06353,REC_06509,REC_06546,REC_06687,REC_05404,REC_04385,REC_02528,REC_03428,REC_07046,REC_01746}} In many imaging studies, multimodality denotes fusion of CT with clinical variables or handcrafted radiomics. Cross-scale integration of imaging with genomics, transcriptomics, or proteomics is much less common and is easily overestimated when molecular markers are merely mentioned. For example, a study predicting EGFR mutation from CT should not automatically be classified as a genomics-input study if genomics is the outcome rather than an input to cancer-risk estimation or existing-lesion malignancy assessment.

A third challenge is that counts of models, modalities, or performance metrics alone do not determine whether these elements form a coherent translational evidence pathway within individual studies. Existing review discussions provide background \cite{REC_05835}; primary modeling studies illustrate the variety of methods to be mapped \reviewblue{\cite{REC_05960,REC_05548,REC_05744,REC_04803,REC_06899,REC_02487,REC_03046}}. These examples motivate the mapping question rather than establish a general shortcoming of all previous reviews. We implement a systematic evidence-mapping strategy that combines conservative bibliographic processing, high-specificity screening, component co-occurrence analysis and a role-aware Multi-Tier Evidence Graph (MTEG). The MTEG separates seven functional tiers: evidence source, patient/context, input modality, representation/model, reasoning/explanation, validation/quality, and clinical output. This role structure prevents conceptually different entities such as radiomics, calibration, and cancer-risk score from being collapsed into one undifferentiated graph.
\reviewgreen{Unlike a PRISMA flow diagram or a conventional evidence table, which primarily records study selection and study-level characteristics, and unlike an ordinary co-occurrence network, which records whether concepts occur together without preserving their translational roles, the MTEG assigns evidence components to predefined functional tiers and evaluates whether those roles are connected within individual studies. Its added value in this review is therefore role-aware reconstruction of the within-study translational pathway and explicit localization of evidence-chain attrition.} 

{The methodology was therefore structured to address these three challenges directly. Full-text task classification separates clinically distinct endpoints that may otherwise share prediction terminology; role-aware modality and omics extraction distinguishes true analytical inputs from outcomes or incidental mentions; and the MTEG organizes extracted evidence into functional tiers so that the continuity of the translational pathway, rather than the presence of isolated components alone, can be evaluated within each study.}

We used five-database retrieval, conservative deduplication, and full-text eligibility review to establish a single corpus for the component, MTEG, and task-space analyses (Fig.~\ref{fig:prisma_flow}).

The principal contributions are an auditable five-database review workflow, mutually exclusive clinical-task classification, role-aware modality/omics extraction, a seven-tier MTEG for evidence-chain analysis, task-specific MTEGs, human eligibility/task review, and structured extraction validation.

\section{Review Questions}
The systematic review and evidence map addressed five questions:
\begin{enumerate}
\item Which clinical, imaging, radiomic, molecular, reasoning, output, and validation components characterize CT/LDCT-based lung-cancer AI?
\item How frequently are these components represented and co-integrated within individual studies?
\item To what extent do studies form complete translational evidence chains spanning patient/context, input modality, model, reasoning/explanation, clinical output, and validation, and where does attrition occur?
\item How is the literature distributed among mutually exclusive Detection, future Risk-prediction, and Other clinical task spaces?
\item How do task-specific MTEGs differ in concept composition, connectivity, and translational evidence architecture?
\end{enumerate}
\section{Methods}
\subsection{Search Strategy and Corpus Assembly}

The search was designed around four mandatory concept blocks: lung cancer/pulmonary nodules, CT/LDCT imaging, AI/ML methodology, and a clinically relevant prediction/detection endpoint. The publication window was 2016--2026. The complete Scopus search strategy was:

\begin{table*}[t]
\centering
\begin{minipage}{0.96\textwidth}
\small
\textbf{Scopus search strategy}

\medskip
{\ttfamily\scriptsize
TITLE-ABS-KEY(
("lung cancer" OR "lung carcinoma" OR "pulmonary cancer" OR
"lung neoplasm*" OR "pulmonary nodule*" OR "lung nodule*")
AND
("computed tomography" OR "low-dose computed tomography" OR
"low-dose CT" OR LDCT OR "chest CT" OR radiomics OR "CT imaging")
AND
("artificial intelligence" OR "machine learning" OR "deep learning" OR
"neural network*" OR "computer vision" OR "large language model*" OR
"vision-language model*" OR multimodal OR "multi-modal")
AND
(predict* OR prognos* OR "risk prediction" OR "risk stratification" OR
"malignancy prediction" OR "malignancy classification" OR diagnos* OR screening)
)
AND PUBYEAR > 2015
AND PUBYEAR < 2027
}
\end{minipage}
\end{table*}

The corresponding database-specific search strategies for PubMed, Embase
(Ovid), IEEE Xplore, and ACM Digital Library are provided in Supplementary
Table~\ref{tab:supp_search_strings}. Supplementary
Tables~\ref{tab:supp_search_strings}--\ref{tab:supp_review_procedures}
additionally report the search date, database-specific retrieval totals,
publication and language rules, deduplication procedure, screening rules,
and full-text eligibility procedures.

Records were retrieved from Scopus (6,963), PubMed (3,184), Embase (4,707), IEEE Xplore (1,774), and ACM (288), yielding \NRaw\ raw records. The final searches of all five databases were conducted on 21 August 2026. The interfaces used were Elsevier Scopus, NCBI PubMed, \reviewblue{Ovid for Embase}, IEEE Xplore Digital Library, and ACM Digital Library.

\subsection{Protocol Registration}
\reviewblue{This systematic evidence map was retrospectively registered in OSF Registries using the Generalized Systematic Review Registration Form on 11 September 2026 (OSF registration: \ReviewRegistration). The review commenced on 1 July 2026, before registration; accordingly, this registration should not be interpreted as prospective preregistration. The registration records the review framework, eligibility criteria, search strategy, extraction strategy, clinical-task classification, evidence-mapping approach, and planned synthesis.}

\subsection{Deduplication and Record Consolidation}
Cross-source deduplication was deliberately conservative, followed by systematic review of the fuzzy-duplicate candidate log. The deduplication procedure was designed to reduce false duplicate removal. Records were linked hierarchically by (i) exact normalized DOI and then (ii) exact normalized title. Title matches were accepted only when the normalized title contained at least 20 characters and there was no conflict in non-empty DOI or publication-year fields. Duplicate relations were consolidated using a Union--Find/disjoint-set structure so that transitive duplicate groups were handled as one component. Within each component, the retained record was the bibliographically richest candidate according to metadata completeness across DOI, title, abstract, authors, journal, year, keywords, and related fields. Ambiguous title matches with DOI/year conflicts were not automatically deleted and were retained for audit. This process reduced the \NRaw\ retrieved records to \NDedup\ consolidated bibliographic records.

\paragraph{Dataset and study-family accounting.}
Bibliographic deduplication identifies duplicate publications but does not by
itself establish independence of the underlying patient cohorts or datasets.
The final analytical denominator therefore represents publications rather than
necessarily independent cohorts. Dataset descriptions and study-family
information were retained at record level using stable \texttt{Record\_ID}
keys; however, a defensible numerical count of unique underlying datasets or
independent study families requires publication-to-cohort identity
adjudication. Automated or normalized dataset-string counts were therefore not
interpreted as counts of independent datasets, institutions, or study
families.

\subsection{Reporting Framework, PICOS-Based Eligibility, and Predefined Criteria}
The review was structured and is reported with reference to PRISMA 2020 \cite{prisma2020} and the eligibility criteria were organized using an adapted PICOS framework. Because the corpus spans diagnostic, screening, prediction-model, segmentation, prognostic, and related AI studies rather than a single intervention design, eligibility was operationalized using a PICOS-based framework adapted to CT/LDCT-centered medical-AI evidence mapping. PICOS was used to make the review question and eligibility boundaries explicit; it was not used to imply that every included study contained an intervention-control comparison. Because the review is an evidence map rather than a pooled intervention-effect review, no quantitative clinical meta-analysis pertaining to it was attempted.

\begin{table*}[t]
\caption{PICOS-Based Eligibility Framework for the CT/LDCT-Centered Lung-Cancer AI Evidence Map}
\label{tab:picos}
\centering
\footnotesize
\begin{tabular}{p{0.10\textwidth}p{0.82\textwidth}}
\toprule
Element & Operational definition\\
\midrule
Population &
Individuals or cohorts evaluated in lung-cancer screening, pulmonary-nodule assessment, lung-cancer diagnosis, future cancer-risk estimation, prognosis, treatment-response assessment, or another prespecified clinically relevant thoracic-oncology setting.\\
Index / intervention &
AI-, machine-learning-, deep-learning-, computer-vision-, or radiomics-based analysis in which CT or LDCT is an analytical input or central imaging component; eligible multimodal extensions could additionally incorporate clinical, longitudinal, molecular, image--text, or other complementary information.\\
Comparator &
Radiologist assessment, pathological or clinical reference standards, conventional risk models, alternative AI/ML models, baseline algorithms, or no explicit comparator when the eligible report was primarily a model-development or validation study.\\
Outcomes &
Future lung-cancer incidence/risk; current lesion or disease detection, segmentation, diagnosis, screening classification, or pulmonary-nodule malignancy; and prespecified related endpoints including prognosis, survival, recurrence, treatment response, staging/progression, lung function, or molecular status.\\
Study design &
Peer-reviewed primary empirical studies reporting development, evaluation, validation, or clinically relevant application of an eligible CT/LDCT-centered computational approach. Reviews, surveys, editorials, dataset-only publications, abstract-only reports, retracted publications, and other non-eligible document types were excluded.\\
\bottomrule
\end{tabular}
\end{table*}

Eligibility criteria operationalize the scope used during corpus assembly and were formalized for transparent reporting of the final analytical mapping. The operational definitions of the three mutually exclusive clinical task spaces are summarized in Table~\ref{tab:taskcriteria}.

Full peer-reviewed conference papers were eligible when they satisfied the same empirical, CT/LDCT, computational, and clinical-endpoint requirements as journal articles. Preprints were excluded as non-peer-reviewed reports, and abstract-only conference records were excluded because they did not provide sufficient full-text methodological evidence for reliable eligibility and extraction. This publication-type rule was applied independently of venue prestige or citation impact.
 Table~\ref{tab:eligibility} separates disease scope, imaging scope, computational methodology, publication type, full-text assessability, language, and evidentiary sufficiency so that exclusions are auditable rather than inferred retrospectively from the final corpus.

\begin{table*}[t]
\caption{Predefined Full-Text Inclusion and Exclusion Criteria}
\label{tab:eligibility}
\centering
\footnotesize
\begin{tabular}{p{0.14\textwidth}p{0.39\textwidth}p{0.39\textwidth}}
\toprule
Domain & Inclusion criterion & Exclusion criterion\\
\midrule
Clinical scope &
Lung cancer, pulmonary nodules, lung-cancer screening, or another directly relevant thoracic-oncology endpoint. &
Primary focus outside lung cancer/pulmonary nodules or an off-target clinical problem without an eligible lung-cancer endpoint.\\
Imaging scope &
CT/LDCT used as a primary analytical input or central imaging component, including eligible multimodal extensions. &
No qualifying CT/LDCT contribution to the analytical model or evidence pathway.\\
Methodological scope &
AI, machine learning, deep learning, computer vision, radiomics, or an eligible multimodal computational model. &
No eligible computational modeling component or purely non-computational analysis.\\
Publication type &
Peer-reviewed primary empirical research with sufficient methodological and outcome information. &
Review/survey, editorial/commentary, dataset-only publication, abstract-only record, retracted article, non-peer-reviewed report, or another non-eligible document type.\\
Full-text availability &
Full text sufficiently accessible to determine eligibility and extract the required evidence. &
Unavailable/inaccessible full text or a document too incomplete for reliable eligibility assessment.\\
Language &
Full text assessable under the review protocol which considers English or in a language that could be reliably assessed by the review team. &
Full text not assessable under the review language protocol.\\
Evidence sufficiency &
Sufficient description of population/data, computational method, and clinically relevant endpoint for evidence mapping. &
Insufficient or inadequately attributable report, invalid/unsafe source, or information insufficient for reliable classification.\\
\bottomrule
\end{tabular}
\end{table*}

\subsection{Independent Human Eligibility and Task Review and Design-Specific Risk-of-Bias Assessment}

Two reviewers independently assessed full-text candidate records for eligibility and classified included studies by clinical task and study design. Initial decisions were recorded before consensus. Disagreements were resolved through discussion, with unresolved cases adjudicated by
the third author. In the original full-text eligibility review, before the final 278-record cohort was established, pre-consensus agreement was 93.5\% (Cohen's $\kappa=0.84$). Design-specific risk-of-bias assessment was conducted separately for all 278 included studies. The dedicated workbook documented a full-text assessment and an independent methodological cross-check by two risk-of-bias reviewers.

Risk-of-bias assessment was design-specific. \reviewblue{Detection studies were assessed using QUADAS-3 v1.2} \cite{quadas3}; \reviewblue{studies in the Risk and Others task spaces were assessed using PROBAST+AI (2025)} \cite{probastai}. \reviewblue{For QUADAS-3, the workbook used Low, High, and II (insufficient information) domain judgments; for PROBAST+AI, it used Low, High, and Unclear.} Domain-level judgments were retained rather than
collapsed into an unvalidated numerical quality score.
\reviewgreen{To examine the relationship between methodological quality and evidence architecture without conflating the two constructs, the locked overall risk-of-bias judgments were descriptively cross-tabulated against primary six-tier chain completeness and the explanation-optional five-tier sensitivity definition. Because QUADAS-3 and PROBAST+AI use partly different judgment labels, these comparisons were descriptive; no pooled numerical quality score or confirmatory cross-tool hypothesis test was applied.} 

Final task labels and risk-of-bias judgments were fixed for analysis.

\subsection{High-Specificity Automated Screening and Full-Text Eligibility}
Screening was designed to prioritize specificity rather than force a prespecified final study count. Illustrative application reports span the screening concepts below \reviewblue{\cite{REC_06122,REC_06199,REC_05483,REC_05743,REC_05543,REC_04360,REC_04211,REC_01417,REC_00136}} Mandatory title/abstract gates required lung-cancer or pulmonary-nodule context, CT/LDCT evidence, AI/ML/deep-learning/radiomics evidence, an eligible clinical or modeling target, and primary empirical-study evidence, with dominant off-target outcomes or modalities excluded. This stage identified \NHighConf\ high-confidence candidates for full-text assessment. The \NBoundary\ near-boundary records, ranked outside the top \NHighConf, were retained separately for the screening audit. None were added to the analytical corpus or contributed to the Detection, Risk-prediction, or Other task-space analyses.

The \NHighConf\ high-confidence candidates underwent retrieval and full-text eligibility review. Of these, \NFullTextExcluded\ were excluded, leaving the \NAnalytical-record analytical corpus. Each excluded record was assigned to one of the exclusion categories listed in Supplementary Table~\ref{tab:fulltext_exclusions}. The exclusion decision was based on eligibility rather than on achieving a target sample size.

\paragraph{Scope of background citations.}
References outside the final analytical corpus are retained for bibliographic context, not as included observations: prior review discussions \cite{REC_05835,REC_05976,REC_03701}, a workshop report \cite{REC_09560}, and other background reports \cite{REC_05853,REC_05508}. The article by Xie and Zhang \cite{REC_04935} was retracted in 2023 (retraction notice doi: 10.1155/2023/9828716); it is outside the 278-study corpus and is cited here solely to document its status. Neither these background references nor the meeting abstract \cite{REC_09382} contributes to the analytical denominators, positive component examples, or co-occurrence estimates.

 The analytical denominator for all downstream component, semantic-network, MTEG, and task-space analyses is therefore \NAnalytical. All final component frequencies, evidence-chain measures, MTEG topology, task-space analyses, and historical benchmarking reported in this manuscript were generated from this frozen \NAnalytical-record analytical corpus; screening-stage candidate counts are used only for review-flow accounting.

\subsection{Full-Text Task Classification and Study Spaces}
Each of the \NAnalytical\ included records was assigned to exactly one task space according to the primary clinical/modeling endpoint actually predicted or analyzed, rather than incidental terminology in the introduction or discussion. The notebook includes explicit assertions that the three primary spaces are mutually exclusive and exhaustive:
\begin{equation}
\mathcal{V}=\mathcal{V}_{D}\cup\mathcal{V}_{R}\cup\mathcal{V}_{O},
\qquad
\mathcal{V}_{D}\cap\mathcal{V}_{R}
=\mathcal{V}_{D}\cap\mathcal{V}_{O}
=\mathcal{V}_{R}\cap\mathcal{V}_{O}
=\varnothing.
\label{eq:taskpartition}
\end{equation}

\begin{table*}[t]
\caption{Operational Criteria for the Three Mutually Exclusive Task Spaces}
\label{tab:taskcriteria}
\centering
\footnotesize
\begin{tabular}{p{0.12\textwidth}p{0.80\textwidth}}
\toprule
Task & Classification criterion\\
\midrule

Risk prediction &
Estimates an individual's susceptibility to, or probability of, developing lung cancer in the future, before an established malignant lesion is being evaluated. This category includes future lung-cancer incidence, susceptibility, individualized risk estimation, and risk stratification. The defining feature is a future lung-cancer diagnosis or incidence endpoint. A nodule visible at baseline may be a predictor when the model estimates cancer diagnosed at follow-up \cite{REC_05652}.\\

Detection &
Identifies, localizes, segments, diagnoses, or characterizes lung cancer or pulmonary lesions that are already present at the time of assessment. This category includes lesion/nodule detection and segmentation, screening detection, diagnostic classification, benign-versus-malignant classification, and malignancy assessment of an existing pulmonary nodule. Accordingly, models estimating ``malignancy probability,'' ``risk of malignancy,'' or the likelihood that an existing nodule is malignant at the index assessment are classified as Detection rather than Risk prediction, despite the use of the term ``risk.''\\

Others &
Includes clinically relevant endpoints that concern neither future occurrence of lung cancer nor detection/characterization of currently present cancer or pulmonary lesions. These include prognosis, overall/progression-free/disease-free survival, recurrence, mortality, treatment response, treatment toxicity or adverse events, cancer staging or progression, pulmonary/lung function, and molecular or mutation-status prediction (e.g., EGFR, ALK, KRAS, or PD-L1), as well as other eligible endpoints outside the preceding two categories.\\

\bottomrule
\end{tabular}
\end{table*}

The derived Non-Risk comparator is Detection plus Others and is used only for matched binary comparison; it is not a fourth primary task class.

\subsection{Role-Aware Omics Detection}
Omics classification was explicitly role-aware. Application reports illustrate the range of molecular endpoints and analytical roles requiring distinction \reviewblue{\cite{REC_05347,REC_06066,REC_04833,REC_04945,REC_05056,REC_05475,REC_04693,REC_03944,REC_07446}} Genomic, transcriptomic, and proteomic terms were counted only when study-specific full-text methods/data/results evidence linked the molecular information to an eligible lung-cancer outcome as an input, feature, predictor, or covariate. Mere presence of terms such as EGFR, ALK, KRAS, TP53, PD-L1, gene expression, or protein markers was insufficient. In particular, CT-to-mutation prediction studies were not labeled as genomics-input studies when mutation status was the prediction target.

Let $O_{ip}$ denote the presence of omics layer $p$ in paper $i$, $R_{ip}$ denote evidence that the molecular layer plays an input/predictor role, and $Y_i$ denote an eligible lung-cancer outcome. The role-aware label was therefore defined conceptually as
\begin{equation}
\widetilde{O}_{ip}=O_{ip}\,\mathbb{I}(R_{ip}=\mathrm{input})\,\mathbb{I}(Y_i\in\mathcal{Y}_{\mathrm{eligible}}).
\end{equation}
This conservative rule was propagated through the component frequency analysis and MTEG. The component-level molecular search was performed specifically within the \reviewblue{278-paper} CT/LDCT-centered AI/radiomics corpus using expanded dictionaries for genetic variants and mutations, gene-expression and RNA/miRNA signatures, protein biomarkers, and related terms. It was not designed to identify all omics-based AI studies in lung cancer; its purpose was to characterize how molecular evidence is integrated into the predefined CT/LDCT-centered corpus.

\subsection{Component Co-Occurrence and Fragmentation}
A paper-component bipartite graph was projected onto the component layer. The projection included clinical, imaging, radiomic, omic, reasoning, validation, and output-related concepts. \reviewblue{Taxonomy coverage was defined as the number of observed modality/component categories divided by the number specified in the frozen taxonomy. Integration density $D$ was the standard undirected density of the projected component co-occurrence graph. Weighted cross-community mixing $M$ was the fraction of total edge weight joining nodes assigned to different detected communities, and the isolated-component fraction $I$ was the proportion of observed component nodes with degree zero.} We report these quantities separately together with modularity and an exploratory bounded fragmentation score
\begin{equation}
F=\frac{(1-D)+(1-M)+I}{3},
\end{equation}
where $D$ is integration density, $M$ is weighted cross-community mixing, and $I$ is the isolated-component fraction. This index is descriptive rather than a validated clinical metric.

\paragraph{Interpretation of component-network fragmentation measures.}
The component-network measures were interpreted separately before considering the composite fragmentation score. Integration density summarizes the breadth of observed connectivity among component concepts in the projected network: larger values indicate that a greater share of potentially connectable component relations is represented, but do not imply that those relations are evenly distributed across conceptual groups. Weighted cross-community mixing quantifies the extent to which observed connection weight spans different detected communities rather than remaining concentrated within recurring within-community combinations. The isolated-component fraction records the proportion of component concepts with no projected co-occurrence connection and therefore captures complete structural separation at the node level. Modularity provides a complementary partition-based description of how strongly the component network separates into internally connected communities. The bounded score $F$ combines lack of integration, lack of cross-community mixing, and isolation into a single exploratory summary; it is interpreted only alongside its constituent measures and is not used as a stand-alone indicator of clinical or methodological quality.

\subsection{Multi-Tier Evidence Graph}
The MTEG contains typed concept nodes $v$ with role
\begin{equation}
r(v)\in\{E,C,I,M,R,V,O\},
\end{equation}
representing evidence source, patient/context, input modality, representation/model, reasoning/explanation, validation/quality, and clinical output, respectively. Paper-to-concept evidence edges and paper-mediated cross-tier co-occurrence edges were retained with provenance.

The patient/context tier explicitly separates screening populations, diagnostic nodule populations, future cancer-incidence risk, nodule malignancy, and prediction horizon. This distinction is essential: predicting future cancer incidence in a screening population is not equivalent to estimating whether an already detected pulmonary nodule is malignant.

The input tier includes clinical variables, radiomics, 2-D/2.5-D/3-D CT, longitudinal CT, genomics, transcriptomics, proteomics, multi-omics, image-text information, and geometric/topological representations. \reviewblue{Input-tier concept detection was restricted to its predefined role-specific input vocabulary; model, reasoning, validation, and clinical-output concepts were assigned to their respective tiers rather than being duplicated as inputs.} \reviewblue{\cite{REC_05515,REC_06116,REC_04961,REC_05053,REC_04352,REC_06948,REC_02467,REC_02988}} The model tier includes classical machine learning, CNNs, vision transformers, multimodal transformers, graph neural networks, foundation-model pretraining, and LLMs. The reasoning tier includes saliency, SHAP, concept-based explanations, ontology, knowledge graphs, first-order logic, rule constraints, and retrieval grounding. Validation includes internal, external, and prospective validation, calibration, decision-curve analysis, clinician evaluation, and factuality or hallucination assessment.
\reviewgreen{Within the MTEG, clinical output denotes the form of model output (for example, diagnosis, malignancy probability, future-risk score, uncertainty estimate, or generated report) and should not be interpreted as clinical actionability. Actionability additionally requires an explicit decision context and supporting evidence such as calibration, decision-curve or utility analysis, external or prospective validation, or clinician evaluation. Consequently, presence of the output tier, or even completion of the evidence chain, is not itself evidence of deployment readiness.} 

A complete evidence chain was defined over six substantive tiers beyond source provenance:
\begin{equation}
C\rightarrow I\rightarrow M\rightarrow R\rightarrow O\rightarrow V.
\label{eq:evidence_chain}
\end{equation}
For paper $p$, six-tier coverage was
\begin{equation}
A_6(p)=\frac{1}{6}\sum_{t\in\{C,I,M,R,O,V\}}\mathbb{I}(p\ \mathrm{covers}\ t).
\end{equation}
A complete chain therefore requires $A_6(p)=1$; the corresponding binary completeness indicator is defined in Table~\ref{tab:mteg_graph_measures}. The broader seven-tier architecture additionally includes evidence-source provenance. The six-tier completeness rule measures within-study co-presence of the required roles. It does not verify a directed clinical reasoning path, causal information flow, or clinical actionability; the schema-oriented graph edges and the coverage test remain distinct constructs.
\reviewgreen{As a sensitivity analysis of the dependence of chain completeness on the reasoning/explanation tier, we additionally evaluated an explanation-optional five-tier chain,
\[
C\rightarrow I\rightarrow M\rightarrow O\rightarrow V,
\]
with $C_5(p)=1$ only when all five of these tiers were present. This sensitivity definition does not replace the primary six-tier MTEG chain; it evaluates whether incomplete integration persists when explicit reasoning/explanation is not mandatory.} 

\subsection{Paper-Semantic Network Construction and Reproducibility}
\reviewblue{The publication-level semantic network was constructed from the frozen 278-study corpus using normalized sentence-transformer embeddings (\texttt{sentence-transformers/all-MiniLM-L6-v2}) and pairwise cosine similarity. The primary undirected weighted paper graph used a deterministic threshold equal to the 92nd percentile of positive pairwise similarities, with a minimum safeguard of 0.25; in the final corpus this yielded a threshold of 0.739336, 278 nodes, and 3,063 edges. An edge therefore denotes semantic similarity at or above the fixed threshold, its weight equals cosine similarity, and its distance attribute is $\max(1-\mathrm{similarity},10^{-9})$. The Detection, Risk, and Others paper networks are induced subnetworks of this same parent graph and therefore use the identical representation and global threshold.}

\reviewgreen{Threshold robustness was examined at similarity quantiles 0.85, 0.88, 0.90, 0.92, 0.94, and 0.96.}

\reviewblue{The MTEG is distinct from the undirected paper-semantic graph. It is a directed, typed evidence graph. Paper-to-concept edges are directed from the study source toward the extracted concept. Paper-mediated concept-to-concept edges are added only for the predefined cross-tier relations (for example, Input modality $\rightarrow$ Representation/model, Representation/model $\rightarrow$ Reasoning/explanation, and Reasoning/explanation $\rightarrow$ Clinical output); no within-tier concept-to-concept co-occurrence edges are introduced by this rule. Repeated support for the same normalized directed concept pair is aggregated as edge weight. Identical normalized concept labels within the same tier collapse to the same concept node. Thus, edge direction in the MTEG represents the predefined functional tier relation, whereas edges in the paper-semantic network are undirected similarities.}

\subsection{MTEG Graph Measures and Evidence-Architecture Interpretation}

MTEG structure was characterized using complementary measures of component
prevalence, connectivity, integration, fragmentation, and evidence-chain
completeness, as defined in Table~\ref{tab:mteg_graph_measures}. These
measures were interpreted descriptively as properties of the evidence
architecture rather than indicators of clinical performance or study quality.

\begin{table*}[t]
\centering
\caption{Core graph-theoretic and evidence-chain measures used to characterize the MTEG evidence architecture.}
\label{tab:mteg_graph_measures}
\scriptsize
\renewcommand{\arraystretch}{1.15}
\setlength{\tabcolsep}{3pt}
\begin{tabularx}{\textwidth}{>{\raggedright\arraybackslash}p{0.17\textwidth} >{\raggedright\arraybackslash}p{0.31\textwidth} X}
\toprule
\textbf{Measure} & \textbf{Definition} & \textbf{Interpretation in evidence synthesis}\\
\midrule
Component prevalence & $P(v)=n_v/N$ & Frequency of each evidence component across studies.\\
Edge frequency & $w_{ij}=\sum_s I_{si}I_{sj}$; $\hat w_{ij}=w_{ij}/N$ & Frequency/proportion with which two components are integrated.\\
Degree & $k_i=\sum_{j\neq i}A_{ij}$ \cite{newman2010} & Breadth of distinct direct component connections.\\
Strength & $s_i=\sum_{j\neq i}w_{ij}$ \cite{barrat2004} & Cumulative recurrence of a component's connections.\\
Betweenness & $BC(v)=\sum_{s\neq v\neq t}\sigma_{st}(v)/\sigma_{st}$ \cite{freeman1977} & Bridging position between evidence regions.\\
Density & $D=2|E|/[|V|(|V|-1)]$ \cite{newman2010} & Proportion of possible pairwise connections observed.\\
Connected components & Number of maximal connected subgraphs & Structural fragmentation.\\
Global efficiency & $E_{\mathrm{glob}}=[|V|(|V|-1)]^{-1}\sum_{i\neq j}1/d_{ij}$ \cite{latora2001} & Network-level accessibility; disconnected pairs contribute zero.\\
Clustering coefficient & $C_i=2e_i/[k_i(k_i-1)]$, $k_i\geq2$ \cite{watts1998} & Local bundling among neighboring evidence components.\\
Complete chain & $C_6(s)=\prod_{t\in\{C,I,M,R,O,V\}}I_{st}$ & Equals 1 only when all six substantive tiers are present.\\
Retention/attrition & $R_k=N_k/N_{k-1}$; $A_k=1-R_k$ & Localizes where successive chains become incomplete.\\
\bottomrule
\end{tabularx}
\end{table*}
\reviewgreen{The explanation-optional five-tier chain $C\rightarrow I\rightarrow M\rightarrow O\rightarrow V$ is evaluated separately as a sensitivity analysis and does not replace the primary six-tier complete-chain definition in Table~\ref{tab:mteg_graph_measures}.}

\subsection{Structured MTEG Extraction Validation}
Extraction validation comprised a full-cohort manual review and a prespecified quantitative audit. The full-cohort workbook covers all \NAnalytical\ eligible \texttt{Record\_ID}s and marks four checks complete: manual full-text validation of tier/concept, model/dataset, and omics-role coding, plus verification of Record ID, title, and full-text alignment. These checks were separate from the human eligibility/task review and design-specific risk-of-bias assessment.

The quantitative audit used a deterministic 100-paper sample frozen on first creation (random state 42). A fixed manifest retained the exact \texttt{Record\_ID}s and audit keys. The audit examined tier assignment and fine-grained concept normalization, model and dataset coding, and omics feature roles. Molecular inputs or predictors were distinguished from outcomes, adjustment variables, stratification factors, and contextual mentions. We evaluated aggregate tiers and fine-grained concepts separately using precision, recall, F1, and Cohen's $\kappa$ where appropriate. The audit assessed extraction reliability and served as a quality-control check.

The component counts, graphs, and six- and five-tier chain summaries use the frozen role-aware extraction labels. The manual annotations provide the validation reference. Molecular frequencies therefore describe the extracted map rather than manually corrected prevalence.

\subsection{Task-Specific MTEG Comparison}
The MTEG taxonomy was projected separately for Detection, Risk prediction, and Others to compare task-specific concept composition and evidence architecture. These graph quantities were interpreted descriptively.

\section{Results}

\subsection{PRISMA, Human Review, and Risk-of-Bias Assessment}
The search, deduplication, screening, full-text eligibility, and
analytical-corpus counts are reported stage-wise in accordance with PRISMA
2020 \cite{prisma2020}. The PICOS-based framework and predefined eligibility
criteria are reported in Tables~\ref{tab:picos} and~\ref{tab:eligibility}.

Risk of bias was assessed for all 278 included studies. QUADAS-3 v1.2 was applied to 216 Detection studies (overall: 88 Low, 70 II, 58 High; applicability: 108 Low, 86 II, 22 High). PROBAST+AI (2025) was applied to 62 Risk/Other studies (overall: 25 Low, 20 High, 17 Unclear; applicability: 36 Low, 26 Unclear). Domain-level judgments are reported in the risk-of-bias workbook and supplementary materials; no aggregate numerical quality score was calculated.
Descriptive linkage of overall risk of bias to the MTEG yielded the following six-tier and explanation-optional five-tier chain completeness rates, respectively: High: 4/78 (5.1\%) and 24/78 (30.8\%); II: 4/70 (5.7\%) and 21/70 (30.0\%); Low: 22/113 (19.5\%) and 60/113 (53.1\%); Unclear: 1/17 (5.9\%) and 5/17 (29.4\%). These descriptive patterns do not support a causal or pooled cross-tool interpretation.

\subsection{Review Accounting, Full-Text Eligibility, Task Classification, and Publication Growth}
\begin{figure}[!t]
\centering
\begin{tikzpicture}[
    font=\scriptsize,
    node distance=3.5mm and 4mm,
    box/.style={
        draw,
        rounded corners=1.5pt,
        align=center,
        minimum height=6.2mm,
        text width=25mm,
        inner sep=2pt,
        line width=0.45pt
    },
    sidebox/.style={
        draw,
        rounded corners=1.5pt,
        align=center,
        minimum height=6.2mm,
        text width=24mm,
        inner sep=2pt,
        line width=0.45pt
    },
    arrow/.style={
        -{Latex[length=1.5mm,width=1mm]},
        line width=0.45pt
    }
]

\node[box] (identified)
{Records identified across five databases\\
$n=16{,}916$};

\node[box, below=of identified] (dedup)
{Records after conservative deduplication\\
$n=9{,}843$};

\node[box, below=of dedup] (automation)
{Records entering strict title/abstract screening\\
$n=9{,}843$};

\node[box, below=of automation] (retained)
{High-confidence full-text candidates\\
$n=400$};

\node[box, below=of retained] (assessed)
{Candidates considered for retrieval and eligibility\\
$n=400$};

\node[box, below=of assessed] (analytical)
{Current full-text analytical corpus\\
$n=\reviewblue{278}$};

\node[box, below=of analytical] (classified)
{Mutually exclusive task classification\\
$n=\reviewblue{278}$};

\node[sidebox, right=6mm of dedup] (duplicates)
{Duplicate records removed\\
$n=7{,}073$};

\node[sidebox, right=6mm of automation] (taexcluded)
{Excluded at title/abstract screening\\
$n=9{,}393$};

\node[sidebox, below=2.5mm of taexcluded] (boundary)
{Near-boundary records retained for separate review\\
$n=50$};

\node[sidebox, right=6mm of assessed] (ftexcluded)
{Retrieval/full-text eligibility exclusions\\
$n=\NFullTextExcluded$};

\node[sidebox, right=6mm of analytical] (boundaryadded)
{Boundary records added to analytical corpus\\
$n=0$};

\node[sidebox, right=6mm of classified] (tasks)
{Detection: $n=\reviewblue{216}$\\
Risk prediction: $n=\reviewblue{7}$\\
Others: $n=\reviewblue{55}$};

\draw[arrow] (identified) -- (dedup);
\draw[arrow] (dedup) -- (automation);
\draw[arrow] (automation) -- (retained);
\draw[arrow] (retained) -- (assessed);
\draw[arrow] (assessed) -- (analytical);
\draw[arrow] (analytical) -- (classified);

\draw[arrow] (dedup.east) -- (duplicates.west);
\draw[arrow] (automation.east) -- (taexcluded.west);
\draw[arrow] ([yshift=-2mm]automation.east) -- ++(3mm,0)
    |- (boundary.west);
\draw[arrow] (assessed.east) -- (ftexcluded.west);
\draw[arrow] (analytical.east) -- (boundaryadded.west);
\draw[arrow] (classified.east) -- (tasks.west);

\end{tikzpicture}

\caption{PRISMA-style accounting of the computational evidence-mapping and full-text eligibility workflow. Five-database retrieval yielded 16,916 records, reduced to 9,843 by conservative DOI/title-based deduplication. Strict title/abstract screening identified 400 high-confidence full-text candidates, 50 near-boundary records retained in a separate audit branch, and 9,393 title/abstract exclusions. Of the 400 high-confidence candidates, 122 were excluded during retrieval and full-text eligibility review, yielding the \reviewblue{278-study analytical corpus}. None of the 50 near-boundary records was added to this corpus. The \reviewblue{278 included analytical studies} were subsequently assigned to mutually exclusive Detection ($n=\reviewblue{216}$), Risk-prediction ($n=\reviewblue{7}$), and Other ($n=\reviewblue{55}$) task spaces.}
\label{fig:prisma_flow}
\end{figure}
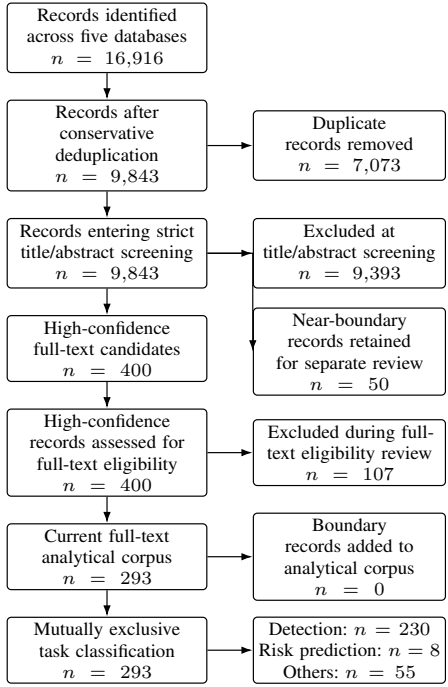

Figure~\ref{fig:prisma_flow} accounts for records from retrieval through task classification, including the separate near-boundary audit branch. Supplementary Table~\ref{tab:prisma_checklist} maps the report to PRISMA 2020 \cite{prisma2020}, and Supplementary Table~\ref{tab:fulltext_exclusions} gives the study-level exclusion categories.

{To provide a conventional overview of the included evidence base, Table~\ref{tab:included_overview} summarizes the clinical-task distribution of the \reviewblue{278 included studies}. This overview complements the subsequent detailed component, semantic-network, and MTEG analyses.}

{The findings are discussed below according to the principal themes of the review. First, we examine the distribution of clinical tasks and methodological components across the included studies. Second, we assess how imaging, clinical, radiomic, and molecular evidence are combined, with particular attention to the interpretation of multimodal inputs. Third, we evaluate whether these components form coherent translational evidence pathways through the MTEG, including reasoning, clinical output, and validation. The accompanying semantic and network analyses provide complementary structural views of these themes.}

\begin{table}[t]
\color{black}
\caption{Overview of the Included Analytical Corpus}
\label{tab:included_overview}
\centering
\scriptsize
\begin{tabular}{lr}
\toprule
Characteristic & Included studies\\
\midrule
Total analytical corpus & \reviewblue{278}\\
Detection & \reviewblue{216 (77.7\%)}\\
Future Risk prediction & \reviewblue{7 (2.5\%)}\\
Other clinical tasks & \reviewblue{55 (19.8\%)}\\
\bottomrule
\end{tabular}
\end{table}

Publication volume rose sharply after 2019, with the largest annual volume in 2025. The apparent decline in 2026 should be interpreted in light of the incomplete calendar year represented in the search export.

\subsection{Data Modalities and Translational Components}
Table~\ref{tab:components} shows the role-aware frequencies computed from the complete study-level mapping, rather than from a citation sample. Illustrative included applications are \reviewblue{\cite{REC_05618,REC_05057,REC_05357,REC_02528,REC_03474,REC_01252,REC_02034,REC_00460}} Clinical variables (\reviewblue{267/278; 96.0\%}), 3-D CT/LDCT (\reviewblue{207/278; 74.5\%}), and radiomics (\reviewblue{178/278; 64.0\%}) dominated the current full-text corpus. External validation was identified in \reviewblue{79 papers (28.4\%)}, calibration in \reviewblue{56 (20.1\%)}, decision-curve analysis in \reviewblue{37 (13.3\%)}, longitudinal CT in \reviewblue{48 (17.3\%)}, and saliency/attribution XAI in \reviewblue{60 (21.6\%)}. Role-aware molecular inputs were mapped less frequently in the fixed extraction: genomics \reviewblue{18 (6.5\%)}, transcriptomics \reviewblue{6 (2.2\%)}, multi-omics \reviewblue{11 (4.0\%)}, and proteomics 1 (0.4\%). The single proteomics-positive pipeline mapping indicates sparse detected proteomic input under the operational extraction definition, rather than omission of proteomics from the taxonomy; because omics-role recall was limited in the manual-reference audit, this is not evidence that only one reviewed study truly used proteomics.

\reviewblue{For descriptive context, we report two-sided 95\% Wilson binomial reference intervals for selected corpus proportions using the 278 included papers as the denominator (Table~\ref{tab:components}). These are descriptive reference intervals for observed paper-level proportions, not inferential confidence intervals for a sampled super-population; they do not account for study-family dependence, heterogeneous designs, or database-selection processes.}

\begin{table*}[t]
\caption{Selected Modalities, Methods, and Translational Components in the \reviewblue{278-Record} Corpus. Citations illustrate one representative study per component; $N$ and percentages are corpus-wide frequencies.}
\label{tab:components}
\centering
\footnotesize
\begin{tabularx}{\textwidth}{lXrrc}
\toprule
Tier/role & Component and representative citation & $N$ & \% & 95\% Wilson reference interval (\%)\\
\midrule
Input & Clinical variables \reviewblue{\cite{REC_06514}} & \reviewblue{267} & \reviewblue{96.0} & \reviewblue{93.1--97.8}\\
Validation & External validation \reviewblue{\cite{REC_05501}} & \reviewblue{79} & \reviewblue{28.4} & \reviewblue{23.4--34.0}\\
Input & 3-D CT/LDCT \reviewblue{\cite{REC_06509}} & \reviewblue{207} & \reviewblue{74.5} & \reviewblue{69.0--79.2}\\
Validation & Calibration \reviewblue{\cite{REC_05321}} & \reviewblue{56} & \reviewblue{20.1} & \reviewblue{15.9--25.3}\\
Input & Radiomics \reviewblue{\cite{REC_05321}} & \reviewblue{178} & \reviewblue{64.0} & \reviewblue{58.2--69.4}\\
Validation & Decision-curve analysis \reviewblue{\cite{REC_05321}} & \reviewblue{37} & \reviewblue{13.3} & \reviewblue{9.8--17.8}\\
Input & 2-D CT/LDCT \reviewblue{\cite{REC_06367}} & \reviewblue{106} & \reviewblue{38.1} & \textemdash\\
Reasoning & Saliency/attribution XAI \reviewblue{\cite{REC_05652}} & \reviewblue{60} & \reviewblue{21.6} & \reviewblue{17.2--26.8}\\
Input & Longitudinal CT \reviewblue{\cite{REC_01746}} & \reviewblue{48} & \reviewblue{17.3} & \reviewblue{13.3--22.1}\\
Reasoning & First-order logic/rules$^{\dagger}$ \reviewblue{\cite{REC_06066}} & \reviewblue{30} & \reviewblue{10.8} & \textemdash\\
Input & Genomics \reviewblue{\cite{REC_01252}} & \reviewblue{18} & \reviewblue{6.5} & \reviewblue{4.1--10.0}\\
Input & Multi-omics \reviewblue{\cite{REC_02528}} & \reviewblue{11} & \reviewblue{4.0} & \reviewblue{2.2--6.9}\\
Input & Transcriptomics \reviewblue{\cite{REC_01150}} & \reviewblue{6} & \reviewblue{2.2} & \reviewblue{1.0--4.6}\\
Input & Proteomics \reviewblue{\cite{REC_02528}} & \reviewblue{1} & \reviewblue{0.4} & \reviewblue{0.1--2.0}\\
Input & 2.5-D imaging \reviewblue{\cite{REC_05475}} & \reviewblue{21} & \reviewblue{7.6} & \textemdash\\
Model & Large language model \reviewblue{\cite{REC_01417}} & \reviewblue{2} & \reviewblue{0.7} & \textemdash\\
Input & Graphs/topology \reviewblue{\cite{REC_03474}} & \reviewblue{7} & \reviewblue{2.5} & \textemdash\\
Output & Generated clinical report \cite{REC_00136} & \reviewblue{2} & \reviewblue{0.7} & \textemdash\\
\bottomrule
\end{tabularx}
\par\vspace{3pt}
\begin{minipage}{0.97\textwidth}
\footnotesize
$N$ is the number of papers mapped to each component by the frozen role-aware extraction pipeline; percentages use \reviewblue{278} as the denominator. Citations identify one representative positive mapping per component, not the sole evidence for its prevalence. A dash denotes an unreported interval. Papers may map to multiple components when multiple roles or modalities are present. $^{\dagger}$``First-order logic/rules'' denotes the taxonomy's combined symbolic/rule-based detector category and does not imply that every mapped study implemented formal first-order logic.
\end{minipage}
\end{table*}

\subsection{Cross-Scale Integration Was Sparse}
\begin{figure*}[t]
\centering
\includegraphics[width=0.94\textwidth]{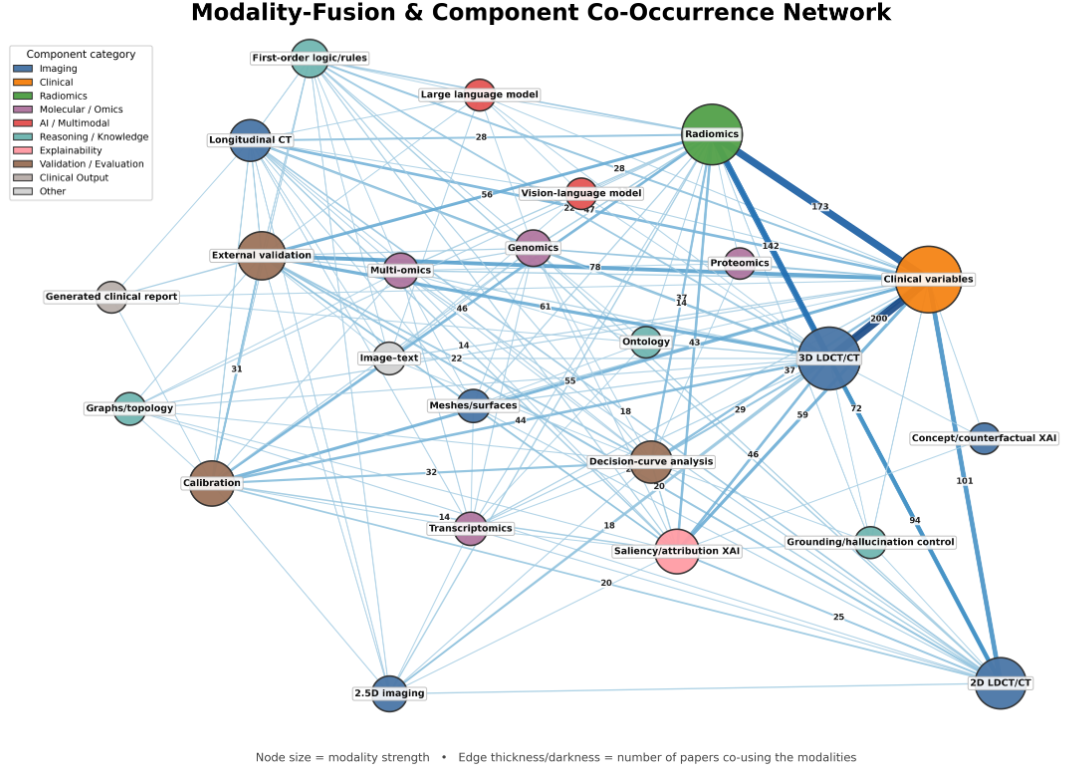}
\caption{Projected component co-occurrence network. Node size reflects prevalence; edge weight reflects paper-level co-use. Omics concepts are visible as a small, weakly connected molecular subnetwork.}
\label{fig:fusion}
\end{figure*}

As visualized in Fig.~\ref{fig:fusion}, the strongest component co-occurrences were 3-D CT/LDCT with clinical variables (\reviewblue{200 papers}), radiomics with clinical variables (\reviewblue{173}), and 3-D CT/LDCT with radiomics (\reviewblue{142}), followed by 2-D CT/LDCT with clinical variables (\reviewblue{101}) and 3-D with 2-D CT/LDCT (\reviewblue{92}) \reviewblue{\cite{REC_05005,REC_05088,REC_04528,REC_04950,REC_02390}}. Molecular integration was much smaller in absolute terms: genomics co-occurred with clinical variables in \reviewblue{18} papers, radiomics in \reviewblue{13}, and 3-D CT/LDCT in \reviewblue{14}; transcriptomics co-occurred with clinical variables in \reviewblue{6} and with radiomics in \reviewblue{5}; multi-omics co-occurred with clinical variables in \reviewblue{11} and radiomics in \reviewblue{9} \reviewblue{\cite{REC_02528,REC_01150,REC_01911}}. The single proteomics-positive paper, \reviewblue{\texttt{REC\_02528}}, co-occurred with clinical variables, radiomics, genomics, and multi-omics \reviewblue{\cite{REC_02528}}. These extraction patterns concentrate mapped integration within the clinical--imaging--radiomics axis, with molecular layers forming a smaller detected cross-scale extension.

The observed taxonomy coverage was \TaxonomyCoverage, component integration density \IntegrationDensity, cross-community mixing \CrossCommunityMixing, and exploratory fragmentation score \FragmentationScore. No observed component node was isolated, but low cross-community mixing indicates that connection was concentrated within a small number of recurring combinations.

\subsection{The MTEG Exposed an Evidence-Chain Bottleneck}
The updated role-aware MTEG contained \NMTEGNodes\ nodes and \NMTEGEdges\ edges. Cross-tier weighted connectivity was strongly asymmetric. Input-to-model links had total weight \reviewblue{1,446} and input-to-clinical-output links \reviewblue{965}, while model-to-reasoning weight was \reviewblue{150} and reasoning-to-clinical-output weight \reviewblue{92}. Validation-to-output weight was \reviewblue{356}. This pattern indicates a much stronger literature on prediction and output generation than on explicit reasoning and validation integrated into the same evidence path.

\begin{sidewaysfigure*}[p]
\centering
\includegraphics[width=0.97\textwidth]{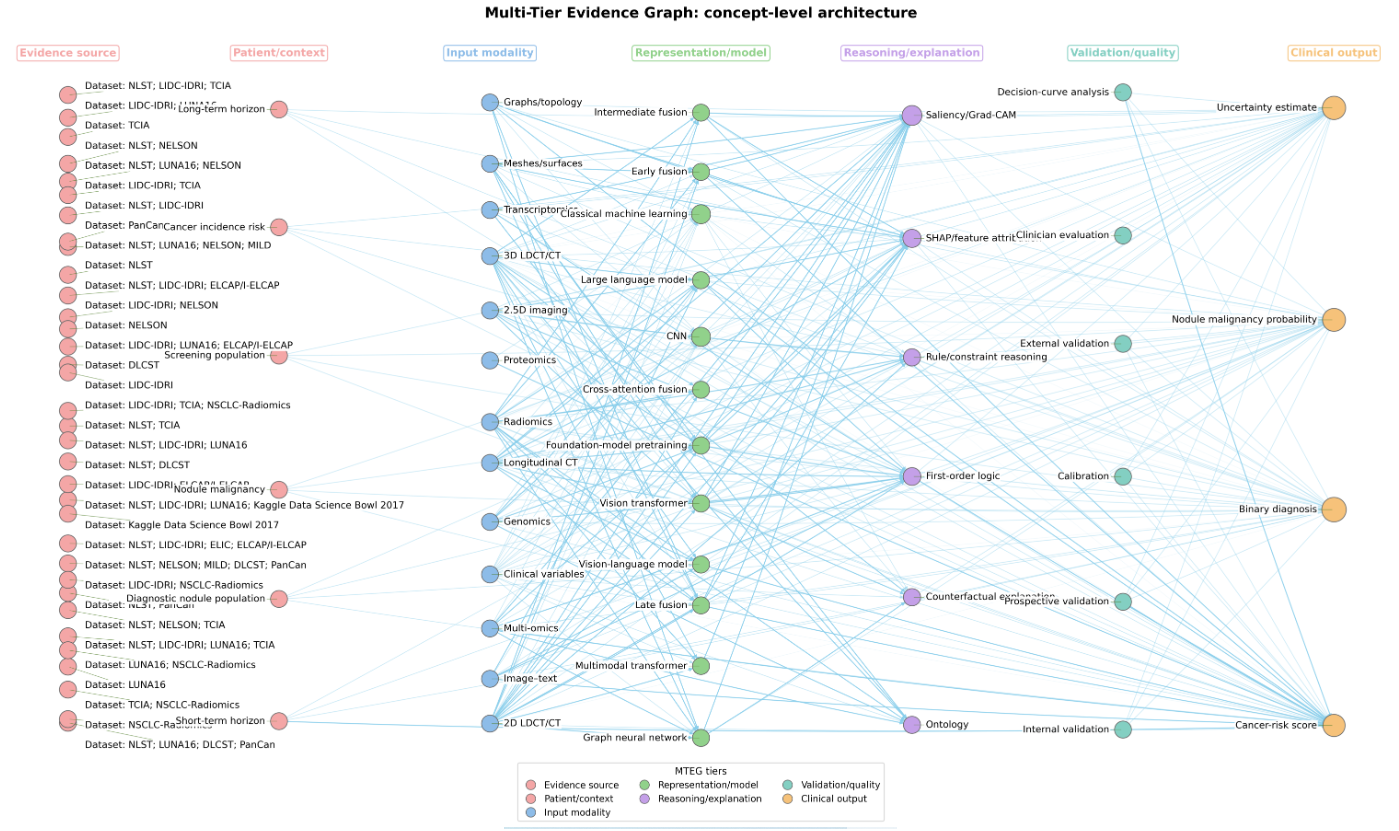}
\caption{Role-aware Multi-Tier Evidence Graph (MTEG) of the lung-cancer CT/LDCT AI evidence architecture. Concepts are arranged from left to right by functional role: Evidence source identifies study provenance; Patient/context defines the clinical population, setting, prediction target, and horizon; Input modality represents clinical, imaging, radiomic, longitudinal, molecular, and multimodal information supplied to models; Representation/model captures machine-learning and representation-learning approaches; Reasoning/explanation includes attribution, explainability, rule-based, ontology, knowledge-graph, and grounding mechanisms; Clinical output represents decision-relevant predictions such as diagnosis, malignancy probability, cancer-risk score, and uncertainty; and Validation/quality captures internal/external validation, calibration, decision analysis, prospective evaluation, and related reliability evidence. Node size reflects PageRank-based structural prominence. \reviewblue{Directed edges denote paper-supported relationships oriented according to the predefined functional cross-tier relation map.} Edge opacity is reduced in highly congested regions and increased in sparsely connected regions solely to improve visual readability and should not be interpreted as a quantitative measure of evidence strength. The graph therefore visualizes how methodological components connect across the translational pathway and where evidence chains remain incomplete.}
\label{fig:mteg}
\end{sidewaysfigure*}

\begin{figure*}[t]
\centering
\includegraphics[width=0.75\textwidth]{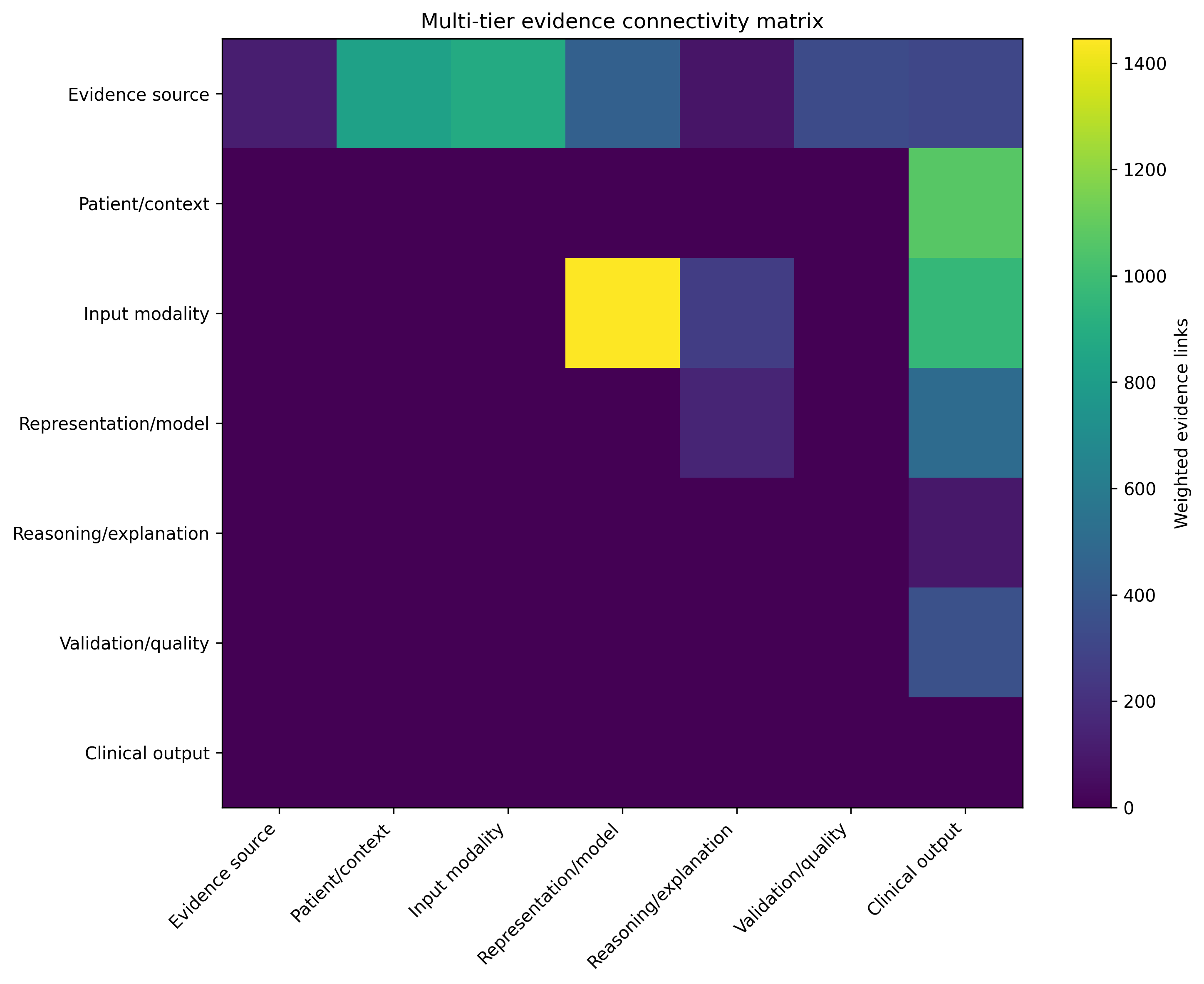}
\caption{Weighted MTEG tier-connectivity matrix. Strong input-to-model and input-to-output links contrast with sparse reasoning-related pathways.}
\label{fig:tiermatrix}
\end{figure*}

The overall role-aware MTEG is shown in Fig.~\ref{fig:mteg}, while the weighted connectivity among its tiers is summarized in Fig.~\ref{fig:tiermatrix}. \reviewblue{Thirty-one of 278 papers (11.15\%)} contained all six substantive chain tiers beyond evidence-source provenance, and the median paper covered \reviewblue{4.5 of six} (Table~\ref{tab:mteg_network_results_summary}). The six-tier chain deliberately excludes source provenance from the completeness criterion; source provenance remains represented in the broader seven-tier graph. Thus the central result is not that components are absent individually, but that end-to-end integration is rare.

\subsection{Central Concepts and Translational Bottlenecks}
Frequency alone does not identify which concepts are structurally prominent in the evidence architecture. Representative high-centrality concepts are summarized in Table~\ref{tab:centrality}. CNN and classical machine learning were supported by \reviewblue{200 and 157 papers}, with degree centralities of \reviewblue{0.263 and 0.263} and betweenness centralities of \reviewblue{0.010 and 0.005}, respectively. These measures describe direct connectivity and bridging within the concept graph and should not be interpreted as clinical superiority.

\begin{table}[t]
\caption{Selected central concepts in the updated role-aware MTEG. Supporting-paper counts and retained centrality measures are taken from the frozen \reviewblue{278-paper} MTEG export.}
\label{tab:centrality}
\centering
\scriptsize
\begin{tabular}{lrrr}
\toprule
Concept & Papers & Degree & Between.\\
\midrule
Classical ML & \reviewblue{157} & \reviewblue{0.263} & \reviewblue{0.005}\\
CNN & \reviewblue{200} & \reviewblue{0.263} & \reviewblue{0.010}\\
Binary diagnosis & \reviewblue{104} & \reviewblue{0.412} & \reviewblue{0.000}\\
Cancer-risk score & \reviewblue{56} & \reviewblue{0.450} & \reviewblue{0.000}\\
Uncertainty estimate & \reviewblue{83} & \reviewblue{0.412} & \reviewblue{0.000}\\
Nodule malignancy prob. & \reviewblue{66} & \reviewblue{0.425} & \reviewblue{0.000}\\
Saliency/Grad-CAM & \reviewblue{27} & \reviewblue{0.287} & \reviewblue{0.002}\\
\bottomrule
\end{tabular}
\end{table}

These centralities are descriptive properties of the concept graph, not measures of clinical superiority or causal importance. They are most useful when interpreted jointly with prevalence, cross-tier connectivity, and evidence-chain completeness.

\begin{table*}[t]
\centering
\caption{Summary of principal findings in the final 278-study corpus and their interpretation for biomedical-AI evidence synthesis.}
\label{tab:mteg_network_results_summary}
\scriptsize
\renewcommand{\arraystretch}{1.16}
\setlength{\tabcolsep}{3pt}
\begin{tabularx}{\textwidth}{>{\raggedright\arraybackslash}p{0.18\textwidth} >{\raggedright\arraybackslash}p{0.29\textwidth} X}
\toprule
\textbf{Network aspect} & \textbf{Observed result} & \textbf{Biomedical/AI interpretation}\\
\midrule
Paper-level semantic network & \reviewblue{278 papers; 3,063 edges; density 0.080} & The literature is selectively interconnected rather than uniformly similar; most possible paper--paper relationships are absent.\\
Connected components & \reviewblue{56 connected components} & The evidence architecture contains multiple structurally separated regions.\\
Local organization & \reviewblue{Mean clustering coefficient 0.342094} & Similar papers tend to form local methodological neighborhoods, consistent with recurring combinations of related approaches.\\
MTEG evidence architecture & \reviewblue{278 mapped papers; 359 nodes; 3,284 edges} & The extracted evidence forms a large relational architecture linking clinical context, inputs, models, reasoning/explanation, outputs, and validation components.\\
Evidence-chain completeness & \reviewblue{31/278 complete chains (11.15\%); median 4.5 tiers} & Most studies cover several translational elements, but only about one in ten connects all six substantive tiers end-to-end.\\
Largest chain bottleneck & \reviewblue{Model $\rightarrow$ reasoning/explanation: 238 $\rightarrow$ 59 studies; conditional retention 24.8\%} & Explicit reasoning/explanation is the sharpest break in the cumulative translational chain, despite broad representation/model coverage.\\
Central representation concepts & \reviewblue{CNN: degree 0.263, betweenness 0.010; classical ML: degree 0.263, betweenness 0.005} & CNN and classical ML have equal reported degree centrality; CNN has higher reported betweenness.\\
Task-space structure & \reviewblue{Detection $N=216$; Risk $N=7$; Others $N=55$} & The literature is dominated by Detection, while genuine future Risk prediction constitutes a small evidence subspace.\\
Fragmentation/integration & \reviewblue{Taxonomy coverage 85.7\%; integration density 0.565; cross-community mixing 0.277; bounded fragmentation 0.386} & Broad conceptual coverage coexists with limited cross-community integration, indicating that evidence breadth does not necessarily translate into end-to-end integration.\\
\bottomrule
\end{tabularx}
\end{table*}

\subsection{Role Awareness Prevented Systematic Omics Inflation}
A keyword-only molecular audit produced substantially more molecular-input mappings than the role-aware extraction definition. { Genomics increased from \reviewblue{18} role-aware positives to \reviewblue{29} keyword-only positives (\reviewblue{1.61-fold}), transcriptomics from \reviewblue{6 to 8} (\reviewblue{1.33-fold}), multi-omics from \reviewblue{11 to 13} (\reviewblue{1.18-fold}), and proteomics from \reviewblue{1} qualifying input study to \reviewblue{four} keyword hits (\reviewblue{4-fold}). } Thus, molecular terminology alone is not a reliable proxy for molecular evidence used as a predictor; role assignment materially changes the inferred evidence architecture. This comparison concerns two extraction definitions and does not establish manual-reference sensitivity: role restriction can reduce terminology-driven positives while still missing true molecular inputs.

\subsection{Structured Extraction Audit Characterized MTEG Reliability}
In the frozen 100-paper audit, macro-F1 was 0.928 for aggregate tiers, 0.683 for fine-grained concepts, and 0.507 for omics-input roles. Omics-input-role macro-recall was 0.380. Table~\ref{tab:tiervalidation} shows variation across tiers. Reasoning/explanation and validation/quality showed strong agreement (F1 0.959 and 0.907, respectively, and Cohen's $\kappa$ 0.935 and 0.632), whereas patient/context and clinical-output labels had lower $\kappa$ despite relatively high recall.

\begin{table}[t]
\caption{Frozen 100-Paper Tier-Level MTEG Extraction Audit. Automated tier assignments are compared with the frozen reference full-text coding used for the structured extraction audit; this audit is distinct from the completed human eligibility/task/risk-of-bias review.}
\label{tab:tiervalidation}
\centering
\scriptsize
\begin{tabular}{lrrrr}
\toprule
Tier & Precision & Recall & F1 & $\kappa$\\
\midrule
Patient/context & \reviewblue{0.874} & \reviewblue{0.954} & \reviewblue{0.912} & \reviewblue{0.042}\\
Input modality & \reviewblue{0.950} & \reviewblue{1.000} & \reviewblue{0.974} & \reviewblue{0.000}\\
Representation/model & \reviewblue{1.000} & \reviewblue{0.938} & \reviewblue{0.968} & \reviewblue{0.476}\\
Reasoning/explanation & \reviewblue{1.000} & \reviewblue{0.921} & \reviewblue{0.959} & \reviewblue{0.935}\\
Clinical output & \reviewblue{0.932} & \reviewblue{0.782} & \reviewblue{0.850} & \reviewblue{0.272}\\
Validation/quality & \reviewblue{0.971} & \reviewblue{0.850} & \reviewblue{0.907} & \reviewblue{0.632}\\
\bottomrule
\end{tabular}
\end{table}

\reviewgreen{For transparency, the corresponding frozen 100-paper aggregate-tier confusion counts (TP/FP/FN/TN) were: patient/context 83/12/4/1; input modality 95/5/0/0; representation/model 91/0/6/3; reasoning/explanation 35/0/3/62; clinical output 68/5/19/8; and validation/quality 68/2/12/18. These counts underlie the precision, recall, F1, and Cohen's $\kappa$ values reported above.} 

\subsection{Stage-Wise Evidence-Chain Attrition Localized the Main Bottleneck}
\begin{figure}[t]
\centering
\includegraphics[width=1.0\columnwidth]{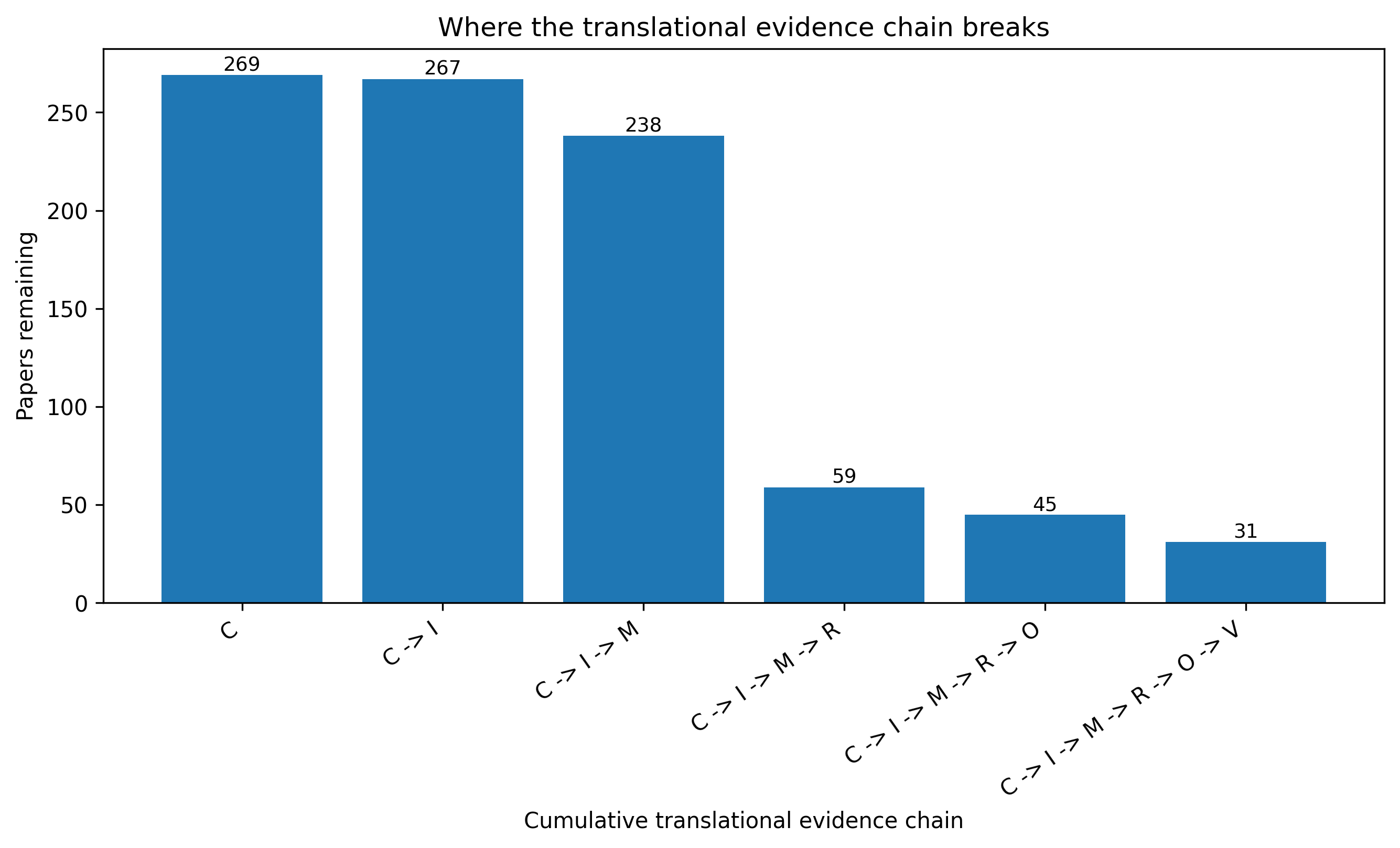}
\caption{Stage-wise attrition across the substantive translational evidence chain. The dominant loss occurs at the transition from representation/model to reasoning/explanation.}
\label{fig:chainattrition}
\end{figure}

Cumulative traversal of the predefined chain revealed where translational integration was lost (Fig.~\ref{fig:chainattrition}). Of the \reviewblue{278} records, \reviewblue{269} contained patient/context evidence and \reviewblue{267} also contained an input modality. Adding a representation/model tier reduced the count to \reviewblue{238}. The largest collapse occurred when reasoning/explanation was required: \reviewblue{59} papers remained, corresponding to \reviewblue{24.8\%} conditional retention from the preceding stage. \reviewblue{Forty-five} papers additionally reached a clinical output and \reviewblue{31} completed the chain through validation.
\reviewgreen{In the explanation-optional sensitivity analysis, cumulative retention across $C\rightarrow I\rightarrow M\rightarrow O\rightarrow V$ was 269, 267, 238, 178, and 110 studies, respectively; thus 110/278 (39.6\%) completed the five-tier chain. Removing the reasoning/explanation requirement therefore increased completeness from 11.15\% to 39.6\%, confirming that explicit reasoning/explanation is a major structural bottleneck while also showing that fewer than half of studies completed the remaining five-tier translational chain.} 

\subsection{Full-Text Classification Revealed Three Clinically Distinct Study Spaces}
Full-text task classification of \reviewblue{278} records produced \reviewblue{216 Detection papers, 7 Risk-prediction papers, and 55 Other applications}. Classification followed the prespecified clinical endpoint boundaries defined in Table~\ref{tab:taskcriteria}, with future lung-cancer occurrence distinguished from detection or characterization of disease already present at the time of assessment. This empirically confirms that broad CT/LDCT ``prediction'' searches retrieve clinically heterogeneous task families rather than a single homogeneous prediction literature.

\reviewblue{Using descriptive two-sided Wilson binomial reference intervals,} Risk prediction represented \reviewblue{7/278 (2.5\%; 95\% CI: 1.2--5.1\%)}, Detection represented \reviewblue{216/278 (77.7\%; 95\% CI: 72.4--82.2\%)}, Other tasks represented \reviewblue{55/278 (19.8\%; 95\% CI: 15.5--24.9\%)}, Non-Risk represented \reviewblue{271/278 (97.5\%; 95\% CI: 94.9--98.8\%)}.


As summarized in Table~\ref{tab:threecomponents}, tier-level coverage showed that all \reviewblue{7} Risk papers contained patient/context, input-modality, and clinical-output evidence, while \reviewblue{4 (57.1\%)} contained a representation/model tier, \reviewblue{2 (28.6\%)} reasoning/explanation, and \reviewblue{4 (57.1\%)} validation/quality. Detection showed \reviewblue{98.1\%} patient/context, \reviewblue{98.6\%} input, \reviewblue{90.7\%} model, \reviewblue{21.8\%} reasoning, \reviewblue{77.8\%} output, \reviewblue{60.2\%} validation coverage. Others showed \reviewblue{90.9\%} patient/context, \reviewblue{100.0\%} input, \reviewblue{83.6\%} model, \reviewblue{32.7\%} reasoning, \reviewblue{47.3\%} output, and \reviewblue{76.4\%} validation coverage.

\begin{table}[t]
\caption{MTEG Tier Coverage Across the Three Full-Text Task Spaces. These are marginal per-tier coverage counts; they should not be confused with cumulative evidence-chain traversal, which requires all preceding tiers.}
\label{tab:threecomponents}
\centering
\scriptsize
\setlength{\tabcolsep}{2.2pt}
\renewcommand{\arraystretch}{1.08}

\begin{tabular}{@{}lrrrrrr@{}}
\toprule
& \multicolumn{2}{c}{Detection}
& \multicolumn{2}{c}{Risk}
& \multicolumn{2}{c}{Others} \\
Tier
& $N$ & \%
& $N$ & \%
& $N$ & \% \\
\midrule

Patient/context
& \reviewblue{212} & \reviewblue{98.1}
& \reviewblue{7} & \reviewblue{100.0}
& \reviewblue{50} & \reviewblue{90.9} \\

Input modality
& \reviewblue{213} & \reviewblue{98.6}
& \reviewblue{7} & \reviewblue{100.0}
& \reviewblue{55} & \reviewblue{100.0} \\

Representation/model
& \reviewblue{196} & \reviewblue{90.7}
& \reviewblue{4} & \reviewblue{57.1}
& \reviewblue{46} & \reviewblue{83.6} \\

Reasoning/explanation
& \reviewblue{47} & \reviewblue{21.8}
& \reviewblue{2} & \reviewblue{28.6}
& \reviewblue{18} & \reviewblue{32.7} \\

Clinical output
& \reviewblue{168} & \reviewblue{77.8}
& \reviewblue{7} & \reviewblue{100.0}
& \reviewblue{26} & \reviewblue{47.3} \\

Validation/quality
& \reviewblue{130} & \reviewblue{60.2}
& \reviewblue{4} & \reviewblue{57.1}
& \reviewblue{42} & \reviewblue{76.4} \\

\bottomrule
\multicolumn{7}{@{}l@{}}{\scriptsize
Detection $N=\reviewblue{216}$;
Risk $N=\reviewblue{7}$;
Others $N=\reviewblue{55}$.}
\end{tabular}
\end{table}

\subsection{Task-Specific MTEGs Revealed Different Translational Concept Architectures}
\begin{sidewaysfigure*}[p]
\centering

\includegraphics[
    width=0.94\textheight,
    height=0.99\textwidth,
    keepaspectratio
]{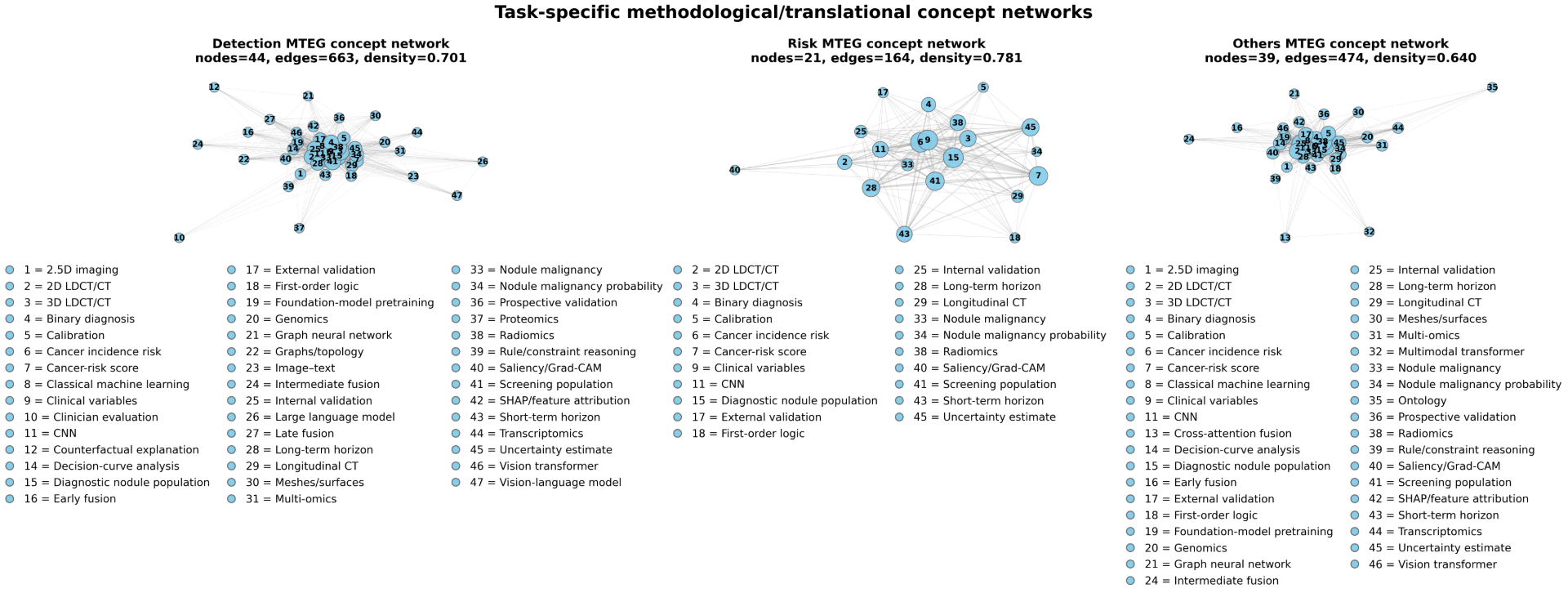}

\caption{
Task-specific MTEG concept networks for Detection, Risk prediction,
and Other applications. Numbered nodes represent methodological and
translational concepts, with the corresponding concept names provided
in the panel legends. Node size reflects weighted concept strength,
while edge width reflects the strength of concept co-occurrence.
Detection uses a \reviewblue{three-column} legend, whereas Risk and Others use
\reviewblue{two-column} legends. These networks describe the organization of methodological and translational concepts rather than the organization of individual papers.
}

\label{fig:threemteg}

\end{sidewaysfigure*}

Figure~\ref{fig:threemteg} shows the task-specific concept networks. The Risk network had the highest density (Table~\ref{tab:mteg3}). Given the differences in graph size and edge construction, density summarizes architectural compactness.

\begin{table}[t]
\caption{Task-Specific MTEG Concept-Network Metrics. Density and clustering refer to the task-specific concept graphs and are therefore distinct from the paper-level semantic-subnetwork metrics in the network comparison.}
\label{tab:mteg3}
\centering
\scriptsize
\begin{tabular}{lrrr}
\toprule
Metric & Detection & Risk & Others\\
\midrule
Papers & \reviewblue{216} & \reviewblue{7} & \reviewblue{55}\\
Concept nodes & \reviewblue{44} & \reviewblue{21} & \reviewblue{39}\\
Concept edges & \reviewblue{663} & \reviewblue{164} & \reviewblue{474}\\
Density & \reviewblue{0.701} & \reviewblue{0.781} & \reviewblue{0.640}\\
Connected components & 1 & 1 & 1\\
Average clustering & \reviewblue{0.053} & \reviewblue{0.273} & \reviewblue{0.077}\\
Edges per paper & \reviewblue{3.07} & \reviewblue{23.43} & \reviewblue{8.62}\\
\bottomrule
\end{tabular}
\end{table}

The Risk MTEG should therefore be interpreted as a smaller but densely interlinked translational concept space rather than a miniature copy of the broader literature. Its higher density and average clustering than the Detection and Others concept networks illustrate why no single topology statistic should be equated with clinical maturity. Component frequency and graph architecture answer different questions: task spaces may use overlapping concepts while arranging them differently within the evidence pathway.

\section{Research Gaps and Evidence Priorities}
The evidence map identifies four principal translational gaps. First, external validation (\PExternal), calibration (\PCalibration), and decision-curve analysis (\PDCA) remain substantially less common than model development. Second, longitudinal CT (\PLongitudinal) is underused despite the temporal nature of future cancer risk. Third, molecular and multi-omic information is infrequently mapped as an analytical input by the fixed role-aware extraction (genomics \PGenomics, transcriptomics \PTranscriptomics, multi-omics \PMultiomics, proteomics \PProteomics). Limited molecular-role recall means these frequencies are provisional mapping estimates rather than exhaustive prevalence. Fourth, explicit reasoning/explanation and clinically grounded output form the largest break in the six-tier evidence chain. Research priorities include clearer separation of clinical endpoints, longitudinal risk modeling, external and prospective validation, calibrated clinical utility, role-aware multimodal integration, and reasoning methods that connect model outputs to clinically interpretable evidence \cite{decideai}. These priorities follow from the observed gaps. Their effect on patient outcomes requires evaluation.

\section{Discussion}

Only 2.5\% of the included studies addressed future lung-cancer
risk. Assessing the malignancy of an existing nodule and estimating
future cancer incidence support different clinical decisions.
Evidence from a detection-dominated literature therefore has
limited applicability to decisions about future risk. The small
number of Risk-prediction studies also limits comparisons of
network density across task groups.

The MTEG identifies which evidence components are reported
together and where coverage drops. Coverage declined most sharply
when reasoning or explanation was required. Even with explanation
optional, fewer than half of the studies covered the remaining
five tiers. The descriptive comparison with risk-of-bias judgments
shows that studies with similar evidence-chain coverage can differ
in methodological quality. Clinical readiness requires evaluation
in the intended setting.

The evidence map can help readers identify gaps in reported
population context, inputs, models, explanations, outputs, and
validation. Readers can verify the underlying full-text evidence
and select clinically comparable studies for task-specific synthesis.
For a candidate system, the intended use should guide further
evaluation, including external or prospective validation,
calibration, decision utility, workflow fit, and safety.

\section{Validation and Generalizability of the MTEG Framework}
The validity of the map rests on the consensus eligibility, task, and study-design decisions, the separate risk-of-bias assessment, and the extraction validation described in Methods. The quantitative audit supports interpretation at the level of broad evidence families and higher-order cross-tier structure, with greater uncertainty for individual normalized concepts.

Downstream robustness analyses addressed different sources of analytical
sensitivity. Similarity-threshold analysis evaluated dependence of semantic
network structure on the primary threshold; bootstrap procedures quantified
stability of selected graph quantities; score ablation tested dependence of
missing-link rankings on individual score components; role-awareness analysis quantified the consequences of collapsing analytical feature roles. Temporal benchmarking was leakage-free,
using only information available before each historical cutoff. Across informative historical cutoffs, the hand-weighted MTEG ranking did not outperform the frequency baseline on average.

The completed validation layers strengthen confidence that the principal
publication-level task distinctions and broad MTEG evidence architecture are
not solely artifacts of automated extraction or graph size. They do not imply
that graph edges are causal, that central concepts are clinically superior, or
that the taxonomy will transfer unchanged to another clinical domain.

External portability also remains unestablished. Replication with a frozen
taxonomy and prespecified analysis protocol in an independent lung-cancer
corpus or another clinical imaging domain would be required to quantify
taxonomy coverage, extraction reproducibility, and stability of the observed
evidence architecture.

\section{Limitations}

The specificity-oriented screening may have missed eligible studies.
Restrictions related to full-text access, language assessability, and
publication type may also have influenced the evidence map. The unit of analysis was the publication, so repeated use of public datasets or overlapping patient cohorts may affect the observed frequencies. The heterogeneity of tasks, endpoints, and study designs limited the synthesis
to descriptive evidence mapping.

Variation in terminology, reporting detail, and the context in which a concept was used may have affected extraction, particularly for fine-grained concepts. The role-aware rule reduced mappings based on molecular keywords alone, but limited recall means that the mapped frequencies cannot establish the prevalence of molecular inputs across the
included studies.

The paper-semantic network depends on the embedding and similarity threshold, while MTEG coverage depends on the extraction and tier definitions. The small future Risk-prediction group ($N=\NRisk$) limits comparisons across task spaces. Graph measures describe the structure of the reported evidence. Establishing clinical effectiveness or readiness for deployment requires evaluation in the intended clinical setting.

\section{Conclusion}

Detection of existing lesions dominates the CT/LDCT lung-cancer AI literature. Only \NRisk\ of \NAnalytical\ studies (2.5\%) addressed future lung-cancer risk. Distinguishing Detection, future Risk prediction, and Other tasks helps identify clinically comparable studies and interpret model performance in the context of the intended clinical decision.

The role-aware MTEG maps how evidence components are reported together within studies and distinguishes molecular inputs from prediction targets and background mentions. All six substantive tiers were covered in \NCompleteChains\ studies, with the greatest drop in
coverage at reasoning and explanation. When explanation was optional, \NCompleteChainsFive\ studies covered the five remaining tiers.

Evidence-chain coverage and risk of bias capture different aspects of the literature. The coverage estimates reflect the frozen automated extraction, whose reliability was assessed against independent manual review. Limited recall for molecular input roles constrains interpretation of the mapped molecular frequencies. The evidence map can guide task-specific synthesis and identify where additional evidence is needed to assess clinical utility.

\section*{Author contributions}
First Author conceptualized and designed the study, developed and implemented the evidence-mapping workflow, curated and analyzed the data, prepared the figures and tables, and drafted the manuscript. Second Author and Third Author conducted the risk-of-bias assessments. All authors read and approved the final manuscript.

\section*{Funding}
This research received no funding.

\section*{Competing Interests}
The authors declare no competing interests.

\setcounter{table}{0}
\renewcommand{\thetable}{S\arabic{table}}

\begin{table*}[t]
\centering
\caption{Master index of supplementary data, review records, validation
materials, and reproducibility artifacts. Public reproducibility materials are available at
\ReviewRepository.}
\label{tab:supp_meta_inventory}

\renewcommand{\arraystretch}{1.15}
\setlength{\tabcolsep}{6pt}
\footnotesize

\begin{tabularx}{\textwidth}{
    >{\raggedright\arraybackslash}X
    >{\raggedright\arraybackslash}X
    >{\raggedright\arraybackslash}X
    >{\raggedright\arraybackslash}X
}
\toprule

\textbf{D1 (All 16916 records)} &
\textbf{D2 (Deduplicated records 9843)} &
\textbf{D3 (Auto excluded record 9393)} &
\textbf{D4 (Eligible Candidates 278)}\\

Database-specific search strategies used for literature retrieval. &
Obtained after the deduplication process. &
Exclusion following low-confidence text assessment. &
Frozen analytical corpus containing the final included studies and stable Record\_ID information used for downstream analyses. \\

\midrule

\textbf{D5 (Ineligible Candidates 122)} &
\textbf{D6 (Backup records 50)} &
\textbf{D7 (MTEG validation 278)} &
\textbf{D8 (Risk of Bias 278)}
\\

Records excluded according to the specified eligibility criteria. &
PRISMA 2020 reporting crosswalk for the systematic evidence map. &
Consolidated MTEG manual validation &
Consolidated Risk of Bias (RoB) report
\\

\bottomrule
\end{tabularx}
\end{table*}

\begin{table*}[p]
\centering
\caption{Database-specific search strategies used for literature retrieval.}
\label{tab:supp_search_strings}

\renewcommand{\arraystretch}{1.15}
\setlength{\tabcolsep}{6pt}

\begin{tabularx}{\textwidth}{
    >{\raggedright\arraybackslash}X
}
\toprule
\textbf{Database-specific search strategy} \\
\midrule

\textbf{PubMed}

\medskip
{\ttfamily\scriptsize
(
"Lung Neoplasms"[Mesh]
OR "lung cancer"[Title/Abstract]
OR "lung carcinoma"[Title/Abstract]
OR "pulmonary cancer"[Title/Abstract]
OR "lung neoplasm*"[Title/Abstract]
OR "pulmonary nodule*"[Title/Abstract]
OR "lung nodule*"[Title/Abstract]
OR "non-small cell lung cancer"[Title/Abstract]
OR NSCLC[Title/Abstract]
OR "small cell lung cancer"[Title/Abstract]
OR SCLC[Title/Abstract]
OR "lung adenocarcinoma"[Title/Abstract]
OR "lung squamous cell carcinoma"[Title/Abstract]
)
AND
(
"Tomography, X-Ray Computed"[Mesh]
OR "computed tomography"[Title/Abstract]
OR "low-dose computed tomography"[Title/Abstract]
OR "low-dose CT"[Title/Abstract]
OR LDCT[Title/Abstract]
OR "chest CT"[Title/Abstract]
OR radiomics[Title/Abstract]
OR "CT imaging"[Title/Abstract]
)
AND
(
"Artificial Intelligence"[Mesh]
OR "artificial intelligence"[Title/Abstract]
OR "machine learning"[Title/Abstract]
OR "deep learning"[Title/Abstract]
OR "neural network*"[Title/Abstract]
OR "computer vision"[Title/Abstract]
OR "computer-aided diagnosis"[Title/Abstract]
OR "large language model*"[Title/Abstract]
OR "vision-language model*"[Title/Abstract]
OR multimodal[Title/Abstract]
OR "multi-modal"[Title/Abstract]
)
AND
(
predict*[Title/Abstract]
OR prognos*[Title/Abstract]
OR "risk prediction"[Title/Abstract]
OR "risk stratification"[Title/Abstract]
OR "malignancy prediction"[Title/Abstract]
OR "malignancy classification"[Title/Abstract]
OR diagnos*[Title/Abstract]
OR screening[Title/Abstract]
)
AND
("2016/01/01"[Date - Publication] : "2026/12/31"[Date - Publication])
}
\\

\addlinespace[10pt]
\midrule

\textbf{IEEE Xplore}

\medskip
{\ttfamily\scriptsize
("All Metadata":"lung cancer"
OR "All Metadata":"lung carcinoma"
OR "All Metadata":"pulmonary cancer"
OR "All Metadata":"lung neoplasm"
OR "All Metadata":"pulmonary nodule"
OR "All Metadata":"lung nodule")
AND
("All Metadata":"computed tomography"
OR "All Metadata":"low-dose computed tomography"
OR "All Metadata":"low-dose CT"
OR "All Metadata":LDCT
OR "All Metadata":"chest CT"
OR "All Metadata":radiomics
OR "All Metadata":"CT imaging")
AND
("All Metadata":"artificial intelligence"
OR "All Metadata":"machine learning"
OR "All Metadata":"deep learning"
OR "All Metadata":"neural network"
OR "All Metadata":"computer vision"
OR "All Metadata":"large language model"
OR "All Metadata":"vision-language model"
OR "All Metadata":multimodal)
AND
("All Metadata":prediction
OR "All Metadata":prognosis
OR "All Metadata":"risk prediction"
OR "All Metadata":"risk stratification"
OR "All Metadata":"malignancy prediction"
OR "All Metadata":"malignancy classification"
OR "All Metadata":diagnosis
OR "All Metadata":screening)
}
\\

\addlinespace[10pt]
\midrule

\textbf{ACM Digital Library}

\medskip
{\ttfamily\scriptsize
[[All: "lung cancer"]
OR [All: "lung carcinoma"]
OR [All: "pulmonary cancer"]
OR [All: "lung neoplasm"]
OR [All: "pulmonary nodule"]
OR [All: "lung nodule"]]
AND
[[All: "computed tomography"]
OR [All: "low-dose computed tomography"]
OR [All: "low-dose ct"]
OR [All: ldct]
OR [All: "chest ct"]
OR [All: radiomics]
OR [All: "ct imaging"]]
AND
[[All: "artificial intelligence"]
OR [All: "machine learning"]
OR [All: "deep learning"]
OR [All: "neural network"]
OR [All: "computer vision"]
OR [All: "large language model"]
OR [All: "vision-language model"]
OR [All: multimodal]]
AND
[[All: prediction]
OR [All: prognosis]
OR [All: "risk prediction"]
OR [All: "risk stratification"]
OR [All: "malignancy prediction"]
OR [All: "malignancy classification"]
OR [All: diagnosis]
OR [All: screening]]
AND
[E-Publication Date: (01/01/2016 TO 12/31/2026)]
}
\\

\addlinespace[10pt]
\midrule

\textbf{Embase/Ovid}

\medskip
{\ttfamily\scriptsize
(
exp lung cancer/ OR
exp lung tumor/ OR
exp non small cell lung cancer/ OR
exp small cell lung cancer/ OR
exp pulmonary nodule/ OR
((lung OR pulmonary) adj3
(cancer\$ OR carcinoma\$ OR neoplasm\$ OR tumor\$ OR tumour\$ OR malignan\$)).ti,ab,kw.
OR (lung adj3 nodule\$).ti,ab,kw.
OR NSCLC.ti,ab,kw.
OR SCLC.ti,ab,kw.
)
AND
(
exp computed tomography/ OR
exp low dose computed tomography/ OR
("computed tomography" OR
"low-dose computed tomography" OR
"low dose computed tomography" OR
"low-dose CT" OR
"low dose CT" OR
LDCT OR
"chest CT" OR
radiomics OR
"CT imaging").ti,ab,kw.
)
AND
(
exp artificial intelligence/ OR
exp machine learning/ OR
exp deep learning/ OR
("artificial intelligence" OR
"machine learning" OR
"deep learning" OR
"neural network\$" OR
"computer vision" OR
"large language model\$" OR
"vision-language model\$" OR
multimodal OR
"multi-modal").ti,ab,kw.
)
AND
(
(predict\$ OR
prognos\$ OR
"risk prediction" OR
"risk stratification" OR
"malignancy prediction" OR
"malignancy classification" OR
diagnos\$ OR
screening).ti,ab,kw.
)
AND
(
2016:2026
).yr.
}
\\

\bottomrule
\end{tabularx}

\end{table*}

\begin{table*}[p]
\centering
\caption{Search, screening, deduplication, and full-text eligibility procedures used to construct the analytical corpus.}
\label{tab:supp_review_procedures}
\renewcommand{\arraystretch}{1.18}
\setlength{\tabcolsep}{5pt}

\begin{tabularx}{\textwidth}{
    >{\raggedright\arraybackslash}p{0.17\textwidth}
    >{\raggedright\arraybackslash}X
    >{\raggedright\arraybackslash}p{0.25\textwidth}
}
\toprule
\textbf{Item} & \textbf{Reported information} & \textbf{Status / qualification} \\
\midrule

\addlinespace

Final search date
&
21-Aug-2026
&
Same date reported in the full manuscript.
\\

\addlinespace

Publication window
&
2016--2026
&
Represented in the query as \texttt{PUBYEAR > 2015} and \texttt{PUBYEAR < 2027}.
\\

\addlinespace

Database-specific retrieved records
&
Scopus: 6,963; PubMed: 3,184; Embase: 4,707; IEEE Xplore: 1,774; ACM: 288; total: 16,916
&
Database-specific search strategies are reported in Supplementary Table~\ref{tab:supp_search_strings}.
\\

\addlinespace

Language rule
&
Full text assessable under the review protocol: English or another language that could be reliably assessed by the review team.
&
Not described as an English-only database search filter.
\\

\addlinespace

Publication-type rule
&
Peer-reviewed primary empirical studies; eligible full conference papers allowed; reviews/surveys, editorials/commentaries, dataset-only publications, abstract-only records, retracted articles, preprints/non-peer-reviewed reports excluded.
&
From the predefined full-text criteria in Table~\ref{tab:eligibility}.
\\

\addlinespace

Deduplication
&
Hierarchical exact normalized DOI followed by exact normalized title; title match required $\geq 20$ normalized characters and no non-empty DOI/year conflict; transitive groups consolidated by Union--Find; bibliographically richest record retained; ambiguous conflicts retained for audit.
&
Conservative deduplication; 16,916 raw records reduced to 9,843.
\\

\addlinespace

Title/abstract screening
&
Mandatory lung-cancer/pulmonary-nodule, CT/LDCT, AI/ML/deep-learning/radiomics, eligible clinical/modeling target, and primary empirical-study gates; dominant off-target outcomes/modalities excluded.
&
Produced 400 high-confidence full-text candidates plus 50 near-boundary records.
\\

\addlinespace

Full-text eligibility
&
Applied the predefined clinical scope, imaging scope, methodological scope, publication type, full-text assessability, language, and evidence-sufficiency rules.
&
\reviewblue{Of 400 high-confidence candidates, 122 were excluded during retrieval and full-text eligibility review, leaving 278 included studies.}
\\

\addlinespace

Full-text exclusion categories
&
Recorded reasons grouped by category, including abstracts, duplicates, insufficient reports, retrieval/access, language notes, clinical or imaging scope, reviews, retraction, dataset-only publications, and other noneligible reports.
&
\reviewblue{The 122 exclusions were categorized using the recorded reasons presented in Supplementary Table~\ref{tab:fulltext_exclusions}.}
\\

\bottomrule
\end{tabularx}
\end{table*}

\begin{table*}[p]
\centering
\caption{Retrieval and full-text eligibility exclusion categories among the 400 high-confidence candidates.}
\label{tab:fulltext_exclusions}
\begin{tabular}{p{0.76\linewidth}r}
\toprule
\textbf{Exclusion category} & \textbf{Count} \\
\midrule
Conference-abstract-only (including poster abstracts) & 22 \\
Residual duplicate & 4 \\
Short report (two pages) with insufficient detail & 10 \\
Non-peer-reviewed report & 3 \\
Unavailable full text & 12 \\
Language-related note & 7 \\
Outside lung-cancer scope & 1 \\
Review/survey & 33 \\
Retracted paper & 4 \\
Dataset-only publication & 2 \\
Pilot/non-eligible document type & 1 \\
Inaccessible through available institutional holdings & 10 \\
Report, news item, or non-research publication & 3 \\
Record not found & 2 \\
Radiography rather than CT/LDCT & 5 \\
Scope conflict (unspecified) & 3 \\
\midrule
\textbf{Total retrieval/full-text eligibility exclusions} & \textbf{122} \\
\bottomrule
\end{tabular}
\par\smallskip
\begin{minipage}{0.96\textwidth}\footnotesize
Counts refer to 122 unique excluded Record IDs. The final 278-study analytical cohort determined inclusion status. Each excluded record was categorized using \texttt{Sel\_Comments} after a leading \texttt{r-} was removed. Language-related comments are retained as recorded labels and do not establish exclusion by study country. The meeting abstract \texttt{REC\_09382} is counted in the conference-abstract-only category.
\end{minipage}
\end{table*}


\begin{table*}[p]
\caption{PRISMA 2020 Reporting Crosswalk for This Systematic Evidence Map}
\label{tab:prisma_checklist}
\centering
\tiny

\begin{tabular}{c p{0.27\textwidth} p{0.58\textwidth}}
\toprule
Item & Reporting topic & Location/status \\
\midrule

1--4 &
Title, abstract, rationale, objectives &
Title/Abstract, Introduction, and \reviewblue{five Review Questions}. \\

5--7 &
Eligibility, information sources, search &
Tables~\ref{tab:picos}--\ref{tab:eligibility}; five databases; final search 21 August 2026; Fig.~\ref{fig:prisma_flow}. \\

8--10 &
Selection, collection, data items &
Deduplication/screening/full-text Methods; role-aware extraction; Table~\ref{tab:taskcriteria}. \\

11 &
Study risk of bias &
\reviewblue{Completed for all 278 studies: QUADAS-3 v1.2 for 216 Detection studies and PROBAST+AI (2025) for 62 Risk/Other studies.} \\

12--13 &
Effect/synthesis methods &
No pooled clinical effect measure; descriptive prevalence, semantic/network analyses, permutation and matched resampling specified. \\

14--15 &
Reporting bias/certainty &
No pooled publication-bias or GRADE claim; selection and evidence limitations are explicitly reported. \\

16--17 &
Study selection/characteristics &
Fig.~\ref{fig:prisma_flow}, corpus/task/component results. \\

18 &
Risk-of-bias results &
Completed for all 278 included studies; summary and domain-level results are reported in the Results, with study-level judgments retained in the completed risk-of-bias workbook. \\

19--22 &
Individual/synthesis results, bias, certainty &
Results tables/figures, \reviewblue{descriptive Wilson reference intervals}, sensitivity analyses, and explicit limits on certainty. \\

23 &
Discussion &
Synthesis, limitations, and implications. \\

24 &
Registration/protocol &
\reviewblue{Retrospectively registered in OSF Registries on 11 September 2026 (\ReviewRegistration).} \\

25--26 &
Support/conflicts &
Funding and competing-interest declarations are reported in the manuscript submission declarations. \\

27 &
Availability &
\ReviewRepository, subject to source-database licensing. \\

\bottomrule
\end{tabular}

\end{table*}

\end{document}